\documentclass[10pt]{article} %
\usepackage[accepted]{rlc}

\usepackage{amssymb}            %
\usepackage{mathtools}          %
\usepackage{mathrsfs}           %
\mathtoolsset{showonlyrefs}     %
\usepackage{graphicx}           %
\usepackage{subcaption}         %
\usepackage[space]{grffile}     %
\usepackage{url}                %

\usepackage{amsmath}
\usepackage{amsthm}
\usepackage{lipsum}

\usepackage{pgfplots}
\DeclareUnicodeCharacter{2212}{−}
\usepgfplotslibrary{groupplots,dateplot}
\usetikzlibrary{patterns,shapes.arrows}
\pgfplotsset{compat=newest}
\usepackage{tabularx}

\theoremstyle{plain}
\newtheorem{theorem}{Theorem}[section]

\newtheorem{lemma}[theorem]{Lemma}

\theoremstyle{definition}
\newtheorem{definition}[theorem]{Definition}

\theoremstyle{remark}

\usepackage{wrapfig}

\usepackage[textsize=tiny]{todonotes}
\definecolor{cpccolor}{RGB}{31,119,180}
\definecolor{vrnncolor}{RGB}{255,127,14}
\definecolor{borelcolor}{RGB}{214,39,40}
\definecolor{contrabarcolor}{RGB}{44,160,44}
\definecolor{oraclecolor}{RGB}{176,176,176}
\definecolor{blindcolor}{RGB}{148,103,189}

\title{Offline Reinforcement Learning from Datasets with Structured Non-Stationarity}

\author{Johannes Ackermann  \\
    ackermann@ms.k.u-tokyo.ac.jp \\
    The University of Tokyo, RIKEN AIP
    \And
    Takayuki Osa \\
    The University of Tokyo, RIKEN AIP \\
    \And
    Masashi Sugiyama \\
    RIKEN AIP, The University of Tokyo
    }

\begin{document}

\maketitle

\begin{abstract}
Current Reinforcement Learning (RL) is often limited by the large amount of data needed to learn a successful policy.
Offline RL aims to solve this issue by using transitions collected by a different behavior policy.
We address a novel Offline RL problem setting in which, while collecting the dataset, the transition and reward functions gradually change between episodes but stay constant within each episode.
We propose a method based on Contrastive Predictive Coding that identifies this non-stationarity in the offline dataset, accounts for it when training a policy, and predicts it during evaluation.
We analyze our proposed method and show that it performs well in simple continuous control tasks and challenging, high-dimensional locomotion tasks.
We show that our method often achieves the oracle performance and performs better than baselines.
\end{abstract}

\section{Introduction}
A main challenge of Reinforcement Learning (RL) is the large amount of interactions required to learn a proficient policy.
One recently popular way to tackle this challenge is to use Offline Reinforcement Learning \citep{levine_offline_2020}.
In Offline RL we aim to learn a policy from a given dataset of previous transitions generated by a different behavior policy, without needing to interact with the environment further.
This avoids the cost and potential risks of online data collection, allowing us to collect large datasets. 
Consider, for example, a policy being trained to improve the controller of a deployed pick and place robot.
Over shorter time frames, we would not expect wear and tear to have a large effect on the robot:
We can expect our environment to be stationary, i.e., the reward and transition functions should be the same over the course of data collection.
However, if we collect the dataset over a longer time frame wear and tear does occur, leading to nonstationarity which causes an important challenge to real world RL \citep{dulac-arnold_challenges_2019}.
With recent works such as \cite{kalashnikov_mt-opt_2021} training on multiple robots over time-frames of up to 16 months, this challenge is becoming increasingly relevant.

As episodes tend to be short compared to the lifespan of a robot, we can assume that the change in transition and reward functions during each episode is small.
We thus tackle this setting by making the structural assumption of a slowly evolving non-stationarity, that remains fixed during each episode and changes between them, allowing us to formulate the setting as multiple rollouts of a Dynamic-Parameter MDP (DP-MDP) \citep{xie_deep_2021}. 
A DP-MDP is a Hidden-Parameter MDP (HiP-MDP) \citep{doshi-velez_hidden_2013} in which the hidden-parameter (HiP) depends on the previous HiPs.
One way to address our problem setting is to use Bayes-Adaptive RL methods, such as BOReL \citep{dorfman_offline_2021_borel} or ContraBAR \citep{choshen_contrabar_2023}.
These methods learn a policy that optimally identifies and exploits the HiP during the same episode.
Another way to approach our problem setting is to derive an offline variant of a lifelong-learning methods like Lifelong Actor Critic (LILAC) \citep{xie_deep_2021} which trains a Dynamic Variational Autoencoder (VAE) \citep{chung_recurrent_2016VRNN} to learn a model of the reward and transition functions.
However, as we will discuss in detail later, these methods contain techniques (reward relabeling, policy replaying, and hard negative mining) that are not applicable in our setting and struggle in high-dimensional tasks with changing transition functions.

We therefore propose a method that avoids the need for these additional techniques and performs well in high-dimensional tasks by using Contrastive Predicitive Coding (CPC) \citep{oord_representation_2019}.
We show that our method is able to learn a meaningful representation of the HiP, identify it in the dataset, predict it during evaluation and use it to learn an effective policy.
To summarize our contributions, we 1) propose a new offline RL problem setting of learning from a dataset including a structured nonstationarity, 2) address this setting by deriving a method based on CPC that infers the non-stationarity in the dataset and predicts it during evaluation, 3) show that our method outperforms baselines in both simple and high-dimensional continuous control tasks and publish our code and datasets for the community to use.

\begin{figure*}[t]
    \centering
    \includegraphics[width=0.95\linewidth]{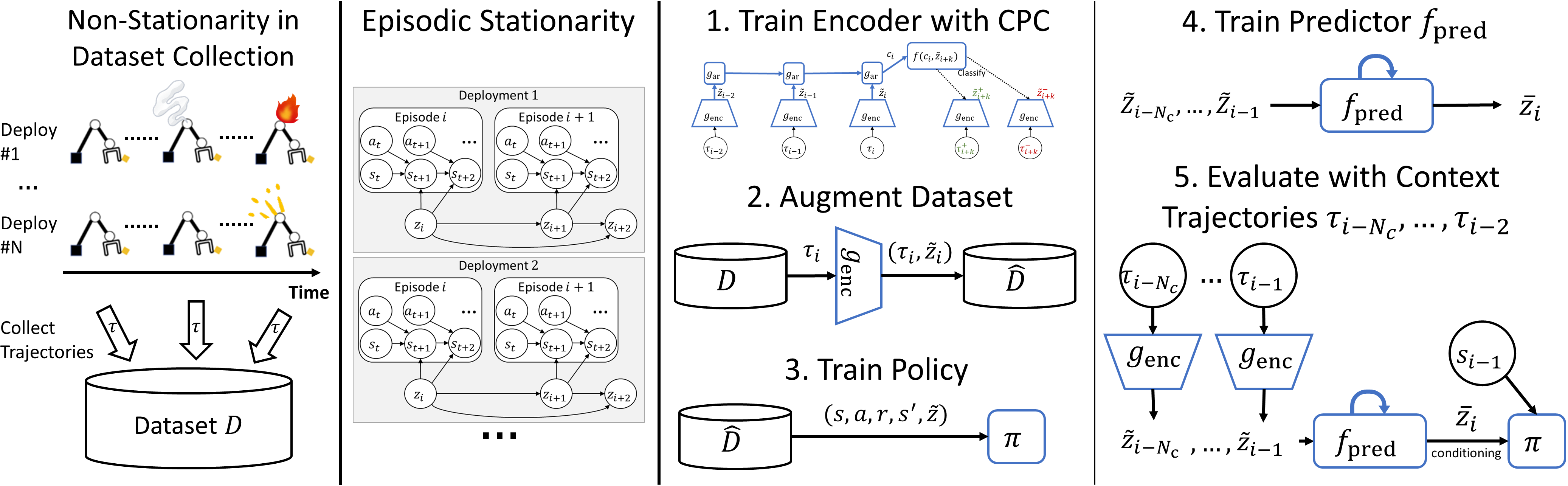}
    \caption{We address an Offline RL setting in which the dataset is generated from multiple deployments with evolving non-stationarity. We make the structural assumption of the reward and transition functions depending on a hidden-parameter $z$ that is constant during each episode but evolves between episodes. Following this assumption, we develop a method based on Contrastive Predictive Coding that infers the hidden parameter from the deployments in our dataset. We then train a predictor and policy to use during evaluation with access to context trajectories.}
    \label{fig:enter-label}
\end{figure*}
\section{Background}
In this section, we will briefly describe the necessary background on RL, CPC and HiP-MDPs.
\subsection{Reinforcement Learning}
In RL, we are given an MDP $M=(S,A,R,p,p_0)$ with, in this work, continuous state space $S$, continuous action space $A$, deterministic reward function $R(s,a)$, transition probability density $p(s'|s,a)$ and initial state density $p_0(s)$ \citep{SuttonRichardS.2002}.
We then aim to learn a policy with conditional probability density $\pi(a|s)$ of choosing action $a$ in state $s$, that maximizes the expected return $J = \mathbb{E}\left[\sum_{t=0}^H r_t\right]$, where $H$ is the duration of an episode.
We refer to a rollout of this MDP as trajectory $\tau=(s_0,a_0,r_0,\dots,s_H,a_H,r_H)$.%
One important quantity of interest is the value function $Q(s,a) = \mathbb{E}[\sum_t \gamma^{t-1} r_t | s_0=s, a_0=a]$.
A common way to estimate it is to use the recurrent formula $Q(s,a) = r(s,a) + \gamma \mathbb{E}[Q(s',a')]$, where $s',a'$ are the subsequent state and action.
In the continuous control setting, Twin Delayed Deep Deterministic Policy Gradient (TD3) \citep{Fujimoto2018} is one popular method to learn the policy.
As an actor-critic method it consists of a critic-network $Q_\theta$ that estimates the action-value function and a deterministic policy $\mu_\phi: S \to A$.
Both are represented as Multi Layer Preceptrons (MLPs) with parameters $\theta$ and $\phi$ respectively and the policy is updated according to the deterministic policy gradient $\nabla_\phi J(\phi)=\mathbb{E}[\nabla_\phi\mu_\phi(s) \nabla_a Q_\theta(s,a)|_{a=\mu_\phi(s)}]$.

In Offline RL, we train a policy from a dataset of transitions $D=\{(s^\mathfrak{i},a^\mathfrak{i},r^\mathfrak{i},s'^\mathfrak{i})\}^{N_\mathrm{D}}_{\mathfrak{i}=1}$ obtained by executing a behavior policy $\beta(a|s)$.
The main challenge in Offline RL is the distribution shift between the states visited and actions chosen by the behavior policy $\beta$ and the learned policy $\pi$ \citep{levine_offline_2020}.
If the shift is large, estimates of the $Q$-value become inaccurate, leading to the policy choosing actions that result in a poor performance.
Therefore, methods such as TD3 do not perform well if applied directly in Offline RL \citep{fujimoto_minimalist_2021}, and instead most methods introduce some way to constrain the learned policy $\pi$ to remain close to the behavior policy $\beta$.
TD3+BC \citep{fujimoto_minimalist_2021} is a successful Offline RL method that achieves this by adding a behavior cloning (BC) penalty $\mathbb{E}_{(s,a) \sim D}[(\pi(s) - a)^2]$ to the policy loss, weighted by a hyperparameter $\lambda >0$.

\subsection{Contrastive Predictive Coding}
CPC \citep{oord_representation_2019} uses contrastive learning to learn a representation $c_t$ for a sequence of observations $o_{1:t}$\footnote{We sometimes use the notation $x_{a:b}:=(x_a,x_{a+1},\dots,x_b)$ for brevity.}.
Each observation $o$ is first encoded by the same encoder $g_{\mathrm{enc}}$, yielding encodings $x_t=g_{\mathrm{enc}}(o_t)$.
These encodings are then processed by an auto-regressive model $g_{\mathrm{ar}}$, such as a Gated-Recurrent Unit (GRU) \citep{Cho2014}, which gives us a representation of the sequence $c_t=g_{\mathrm{ar}}(x_{1:t})$.
We then take the future observation $o^+_{t+k}$ from the same sequence as a positive sample and sample a set of $N^-$ observations $\{o^{-,j}_{t+k}\}_{j=1}^{N^-}$ from different sequences as negative samples.
A classifier $f_k(c_t, x_{t+k})$ is then trained to classify which embeddings are from the positive sample $x_{t+k}^+=g_{\mathrm{enc}}(o_{t+k}^+)$ or negative samples $x_{t+k}^{-,j}=g_{\mathrm{enc}}(o_{t+k}^{-,j})$.
The encoder $g_{\mathrm{enc}}$, autoregressive model $g_{\mathrm{ar}}$ and classifier $f$ are trained jointly to optimize the InfoNCE loss:
\begin{equation}
\label{eq:infonce}
\begin{aligned}
    \mathcal{L}_{\mathtt{InfoNCE}}=  
    -\mathbb{E}\left[\log\frac{\exp f_k(c_t, x_{t+k}^+)}{\exp f_k(c_t,x^{+}_{t+k}) + \sum_{j=1}^{N^-}\exp f_k(c_t, x_{t+k}^{-,j})}\right] \,,
\end{aligned}
\end{equation}
which is minimized when the classifier is proportional to the density ratio of a sample being from the conditional density $p(o_{k+t}|c_t)$ instead of the proposal density $p(o_{k+t})$, i.e., $f(o_{t+k},c_t) \propto \frac{p(o_{k+t}|c_t)}{p(o_{k+t})}$, thereby maximizing the mutual information $I(c_t;o_{t+k})$ \citep{oord_representation_2019}.

\subsection{Partially-Observable MDPs}
In the classical MDP the reward function $R$ and transition function $P$ are stationary during training, i.e., they do not change.
While this assumption is easily satisfied in constructed examples, in realistic settings external influences and unobservable factors can make it difficult or impossible to choose a state formulation that permits a stationary transition and reward functions.
Partially-Observable MDPs (POMDPs) \citep{astrom_optimal_1965} extend the MDP formulation by assuming that while the transition and reward functions seem non-stationary from the given observation $s$, they are stationary given the unobserved state $\hat{s}$. The observation is given by an observation function $h: \hat{S} \to S$.
We can represent any kind of non-stationary transition- or reward-function as a POMDP, however, the generality of the formulation makes efficient training difficult.
HiP-MDPs introduced by \citet{doshi-velez_hidden_2013} address this issue by constraining the true state $\hat{s}$ to be the combination of the observation received by the agent with a hidden parameter $z$, i.e. $\hat{s} = (s,z)$. 
While in general continuous sets of HiPs can be considered, following related works \citep{xie_deep_2021,dorfman_offline_2021_borel} we focus on a discrete set of HiPs $Z$ in our experiments and thus also in the rest of this work.
This HiP is sampled at the beginning of each episode from a distribution $z\sim P(z)$ and remains constant throughout the episode.
The transition function $P(s'|s,a,z)$ then depends on the hidden parameter $z$.
DP-MDPs \citep{xie_deep_2021}, visualized in Fig. \ref{fig:dpmpg-and-cpcapproach} (left), generalize the HiP-MDP by considering a structured evolution of HiPs:
\definition{\textbf{Dynamic-Parameter MDP} \citep{xie_deep_2021}}{ A DP-MDP is a is an MDP with the addition of a HiP space $Z$, transition probability $P_\mathrm{z}(z_i|z_{0:i-1})$, and initial probability $P_\mathrm{z_0}(z_0)$. The HiP is constant during each episode and follows $P_\mathrm{z}(z_i|z_{0:i-1})$ between episodes. The transition density $p(s'|s,a,z)$ and reward function $R(s,a,z)$ depend on the HiP.}

\subsection{Bayes-Adaptive RL}
The HiP-MDP setting can be addressed by Bayes-Adaptive RL methods, which train a policy that infers and exploits the HiP during an episode.
Variational Bayes Adaptive Deep RL (VariBAD) \citep{Zintgraf2020} achieved this by training a VAE with a Gaussian encoder $[\mu_t , \Sigma_t] = q_\phi(s_{1:t},a_{1:t},r_{1:t})$, reward decoder $p_{r,\phi}(r_{t'}|s_{t'},a_{t'},\tilde{b}_t)$ and transition decoder $p_{t,\phi}(s'_{t'}|s_{t'},a_{t'},\tilde{b}_t)$, for $1 \leq t'< < t$, where $\tilde{b}_t \sim \mathcal{N}(\mu_t,\Sigma_t)$.
$\mathcal{N}(\mu,\Sigma)$ is a multivariate Gaussian distribution with mean $\mu$ and diagonal covariance matrix $\Sigma$.
$b_t=(\mu,\Sigma)$ is the belief over the current HiP, which the policy is then conditioned on.
\cite{dorfman_offline_2021_borel} investigated the application of VariBAD to offline datasets generated by behavior policies $\beta(a|s,z)$, conditioned on the task HiP $z$, and introduced Bayes Adaptive Offline Reinforcement Learning (BOReL) with two new techniques that enable successful training:
\textit{Reward Relabeling} which for each transition $(s,a,r,s')$ in the dataset creates additional transitions $(s,a,R(s,a,z_i),s')$ for all HiPs $z_i \in Z$. 
\textit{Policy Replaying}, which generates trajectories with HiP $z$ using the behavior policy conditioned on each other HiP $z_i \not=z$. 
\cite{choshen_contrabar_2023} introduced Contrastive Bayes Adaptive Deep RL (ContraBAR) which instead of a VAE uses CPC to learn a belief over the HiP.
Specifically, it encodes transitions using an encoder $z_t = g_\mathrm{enc}(s_t,a_{t-1},r_t)$ and combines them using a Recurrent Neural Network (RNN) to a belief $b_t=g_\mathrm{ar}(z_{1:t})$, which is trained by discriminating the next transitions $(s_{t+k}^+, r_{t+k}^+)$ from the same episode against those from a different episode $(s_{t+k}^-,r_{t+k}^-)$.
However, as they pointed out, it is possible to discriminate the future transition not by learning a belief over the HiP but by learning a transition model $p(s_{t+k}|s_t)$, leading the training to fail.
To prevent this they used \textit{hard negative mining}, either by reward relabeling when only the reward changes or by simulating transitions when the transition function changes, requiring access to a simulator of the environment.
As we will see in the next section, \textit{reward relabeling} is not applicable in our setting as we do not consider access to the reward function, \textit{policy replaying} is not applicable as our behavior policy is not conditioned on $z$ and \textit{hard negative mining} is not applicable as we do not have access to a simulator of the environment.

\subsection{Problem Formulation}
\label{seg:problemformulation}
Recall that our motivation is a setting in which data is collected from multiple deployments over an extended time-frame.
As episodes tend to be short compared to the lifespan of a deployment, it is reasonable to assume stationarity during the duration of each episode, making the DP-MDP \citep{xie_deep_2021} a natural fit.
We further assume that the data is generated by a behavior policy $\beta(a|s)$, that does not have access to the HiP $z$, for example a robust controller that performs well but not optimally on all HiPs. 
We aim to improve on this behavior policy by inferring and using the HiP $z$.

\textbf{Problem Setting:}{ We are given a dataset $D=\{d^j\}_{j=1}^N$ consisting of $N$ deployments $d=(\tau_1,\dots,\tau_i,\dots,\tau_M)$, each containing $M$ trajectories $\tau_i$. 
Each deployment $d$ is a rollout of the same DP-MDP $\mathcal{M}$. 
The deployments are generated by, for each deployment, first sampling a HiP sequence $z_0\sim P_{\mathrm{z}_0}(z)$, $z_i\sim P_\mathrm{z}(z_i|z_{1:i-1})$ and then each trajectory $\tau_i$ is generated by sampling $s_0\sim p_0(s)$ and following behavior policy $\beta(a|s)$, transition density $p(s'|s,a,z_i)$, and reward function $R(s,a,z_i)$.
During evaluation we are given a context of $N_\mathrm{c}$ previous trajectories $\tau_{i-N_\mathrm{c}:i-1}$ generated by the behavior policy $\beta$ and our objective is to learn a policy conditioned on the context $\pi(a|s,\tau_{i-N_c:i-1})$ that maximizes the return over the next episode, i.e.,
$J = \mathbb{E}_{\pi,P_\mathrm{z},P,P_0,z_{i-N_\mathrm{c}:i-1},\tau_{i-N_\mathrm{c}:i-1}}\left[\sum_{t=0}^{H}R(s_t,a_t,z_i) \right] \,.$ 
}

\section{Algorithm}
We introduce our proposed method named Contrastive Predictive Non-Stationarity Adaptation (COSPA).
As the reward and transition functions depend on the HiP $z_i$, we first must infer it in the dataset, use it to train a policy and then predict it during evaluation.
The offline setting allows us to separate these steps and train an inference model and predictor model first. 

\paragraph{Inferring the Hidden Parameter}
\begin{figure}
    \begin{minipage}{0.39\linewidth}
        \centering
        \includegraphics[width=0.9\linewidth]{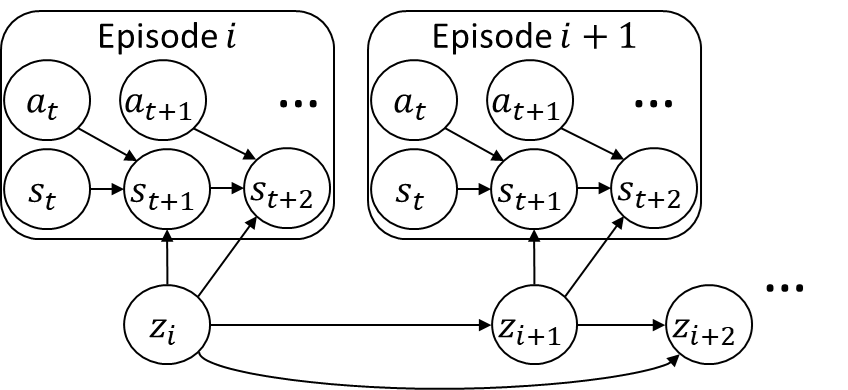}
    \end{minipage}
    \begin{minipage}{0.6\linewidth}
        \centering
        \includegraphics[width=0.95\linewidth]{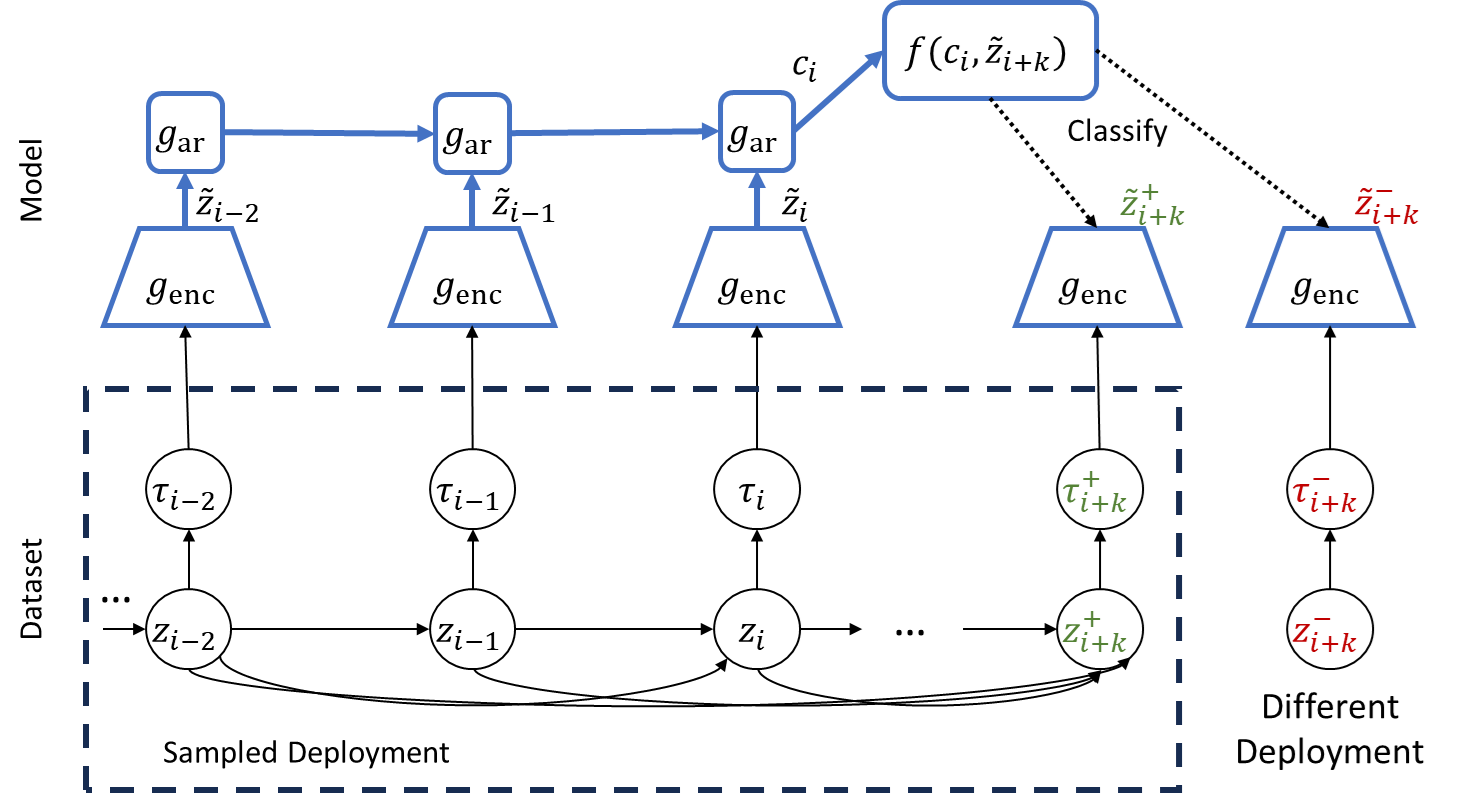}
    \end{minipage}
    \caption{Left: Graphical model of the DP-MDP. Right: Illustration of a deployment sampled from the dataset and our approach to infer the hidden variable. We use Contrastive Predictive Coding to learn a model that can discriminate future trajectories $\tau_{i+k}$ based on past trajectories $(\tau_{i},\tau_{i-1},\dots,\tau_1)$ by learning a representation of the past trajectories $(\tilde{z}_i,\tilde{z}_{i-1},\dots,\tilde{z}_1)$.}
    \label{fig:dpmpg-and-cpcapproach}
\end{figure}
From the generative model of the DP-MDP, shown in Fig. \ref{fig:dpmpg-and-cpcapproach}, we know that the next episode $\tau_i$ with HiP $z_i$ only depends on the HiPs $(z_{i-1},z_{i-2},\dots,z_1)$ of previous episodes $(\tau_{i-1},\tau_{i-2},\dots,\tau_1)$.
One way of learning to infer an approximate HiP $\tilde{z}$ in this setting is to derive an offline variant of LILAC \citep{xie_deep_2021}. LILAC trains a Dynamic VAE with an encoder $\tilde{z}_i = q_\phi(\tau_i)$, dynamic prior $p_\psi(\tilde{z}_i|\tilde{z}_{i-1},\dots,\tilde{z}_1)$ and decoder $p_\phi(\tau_i|\tilde{z}_i)$.
While training an accurate decoder $p_\phi(\tau_i|\tilde{z}_i)$ is feasible in settings with varying reward functions or in low-dimensional problems, it becomes challenging in high-dimensional settings with small variations in the transition function, as we will see in the experiments in Section \ref{sec:evaluation}.

Instead of learning a generative model $p_\phi(\tau_i|\tilde{z}_i)$, it is often easier to learn a discriminative model. %
This makes the application of contrastive learning and in particular CPC a natural choice for our problem setting.
As shown in Fig. \ref{fig:dpmpg-and-cpcapproach} (right), we treat each trajectory $\tau$ as an observation of a time-sequence, encode them separately using an encoder $\tilde{z}_i = g_{\mathrm{enc}}(\tau_i)$ and then summarize the past encodings to a context $c_i$ using an autoregressive encoder $c_i=g_{\mathrm{ar}}(\tilde{z}_i,\tilde{z}_{i-1},\dots,\tilde{z_1})$.
Finally, in the InfoNCE loss, a classifier $f(c_t,\tau_{t+k})$ is used to distinguish a future trajectory $\tau_{i+k}^+$ of the same deployment from negative trajectory samples $\{\tau_{i+k}^{-,j}\}_{j=1}^{N^-}$ from different deployments.
The InfoNCE loss in \eqref{eq:infonce} therefore becomes
\begin{equation}
\label{eq:cpcreprloss}
\begin{aligned}
    \mathcal{L}_{\mathrm{repr}}=  
    -\mathbb{E}\left[\log
    \frac
    {\exp f_k(\tau_{i+k}^+, c_{i})}
    {\exp f_k(\tau_{i+k}^+, c_{i}) + \sum_{j=1}^{N^-}\exp f_k(\tau_{i+k}^{-,j}, c_{i})
    }\right] \,.
\end{aligned}
\end{equation}
This structure is shown in Fig. \ref{fig:dpmpg-and-cpcapproach} (right). 
As we learn to discriminate future trajectories $\tau_{i+k}^+$, instead of future transitions $(s_{t+k}^+,r_{t+k}^+)$ as in ContraBAR, the model can not simply learn the transition function $p(s_{t+k}|s_t)$ but has to learn a representation of the HiP $z$ to discriminate $\tau_{i+k}$.

Following the same argument as \citet{oord_representation_2019}, we can show that minimizing this loss maximizes the mutual information $I(c_i;\tau_{i+k})$, which we show in Appendix \ref{app:analysisofcpc} for completeness.
We could thus directly use $c_i$ as an approximation of the hidden parameter $z_{i+k}$.
However, this would have the disadvantage of requiring at least $k$ episodes as context during evaluation, and prevent usage of the first $k$ episodes per deployment during training.
We avoid this issue by using the output of the encoder $\tilde{z}_i = g_{\mathrm{enc}}(\tau_i)$ as an approximation of the HiP, obtaining the augmented dataset $\hat{D}$ with $\hat{s}=(s,\tilde{z})$ to train the policy $\pi$, and train a separate prediction network to use during evaluation.
\paragraph{Predicting the Next Hidden State}
During evaluation we only have access to a context of $N_\mathrm{c}$ previous trajectories and need to infer the next HiP $\tau_{i+1}$ to condition our policy on.
We therefore train a predictor RNN $f_\mathrm{pred}$ to predict the next HiP $z_{i+1}$.
To train it, we sample sequences of inferred latents $(\tilde{z}_{i-N_\mathrm{c}},\dots,\tilde{z}_{i})$ from the dataset $\hat{D}$ and minimize the mean squared error $\mathcal{L}_\mathrm{pred} = \mathbb{E}_{(\tilde{z}_{i-N_\mathrm{c}},\dots,\tilde{z}_{i})\sim \hat{D}}[(f_\mathrm{pred}(\tilde{z}_{i-N_\mathrm{c}},\dots,\tilde{z}_{i-1}) - \tilde{z}_i)^2]$.
As we will show in Section \ref{sec:evaluation}, we found that this works well even with relatively simple network structure, consisting of two hidden layers and a GRU.

\paragraph{Reinforcement Learning}
Having inferred the HiP and relabeled our offline dataset as $\hat{s}=(s,\tilde{z})$, we now need to train a policy $\pi(a|\hat{s})$.
While in principle any Offline RL method may be used, we need to consider how our problem setting differs from the popular D4RL benchmark \citep{fu_d4rl_2021}, which many popular methods are designed for.
More so than in D4RL, our setting requires large deviations from the behavior policy due to the difference in transition and reward functions. This makes methods such as Advantage-Weighted Regression \citep{peng_advantage-weighted_2019} or Implicit Q Learning \citep{kostrikov_offline_2021} that contain strong constraints to in-dataset actions disadvantageous.
TD3+BC \citep{fujimoto_minimalist_2021} uses a deterministic policy $\mu_\phi(s)$ and conservativity is achieved by a BC term $\mathbb{E}_{(s,a)\sim D}\left[(\mu_\phi(s) - a)^2\right]$.
In our setting we extend this to $\mathbb{E}_{(s,a,\tilde{z})\sim \hat{D}}\left[(\mu_\phi(s,\tilde{z}) - a)^2\right]$, which can be estimated using samples from the augmented dataset $\hat{D}$. 
We can adjust the strength of this constraint by varying the hyperparameter $\lambda$, allowing more flexibility to learn the policy $\mu_\phi$, and thus use TD3+BC in our experiments.

\section{Implementation and Experiments}
We make several design choices in the implementation of our method to enable efficient training.
We implement $g_{\mathrm{enc}}$ as a two-layer MLP with ReLU activations, $g_{\mathrm{ar}}$ as a GRU \citep{Cho2014} and the classifier $f$ as an MLP with two hidden-layers.
One important consideration is how to encode the trajectory $\tau$ with encoder $g_{\mathrm{enc}}$.
To enable efficient training, we use sampled transitions $(s_t,a_t,r_t,s_{t+1})$ as input.
This works well in tasks with changing transition dynamics, such as changes to the configuration of a robot, or dense reward functions. 
For sparse reward tasks recurrent encoders can be considered, but we focus on the case with changing transition functions.
When augmenting the dataset, we average the output over each trajectory, i.e., each transition of trajectory $\tau_i$ is augmented with the average embedding $\frac{1}{H}\sum_{t=1}^{H-1} g_{\mathrm{enc}}(s_t,a_t,r_t,s_{t+1})$ of the trajectory.
We further use a low dimensionality ($2,4,8$) for $\tilde{z}$ and normalize it before using it in the policy training.
The prediction RNN $f_\mathrm{pred}$ is implemented as a two-layer MLP followed by a GRU.
For TD3+BC, we use similar parameters to the ones proposed by the authors \cite{fujimoto_minimalist_2021}, but decrease the strength of the BC penalty to account for the larger difference between behavior policy and optimal policy.
Finally, we also add layer-normalization \citep{ba_layer_2016} after each hidden layer of the critic, as suggested by \citet{kumar_dr3_2022}.
We implement our method using JAX and publish our implementation.\footnote{see \url{https://github.com/JohannesAck/OfflineRLStructuredNonstationarity}} 
We use the same RL hyper-parameters per environment for all methods, but optimize the hyper-parameters of our baseline representation methods by extensive grid-search per task on the low-dimensional and Ant tasks.
Additional details can be found in Appendix \ref{app:implementation_details}.

\subsection{Evaluation}
\label{sec:evaluation}
\begin{figure}
    \begin{tabularx}{\linewidth}{cccccc}
        1D-Goal & 2D-Goal & 2D-Wind & Ant-Leg & Ant-Weight & Barkour-Weight \\
        \includegraphics[height=3cm]{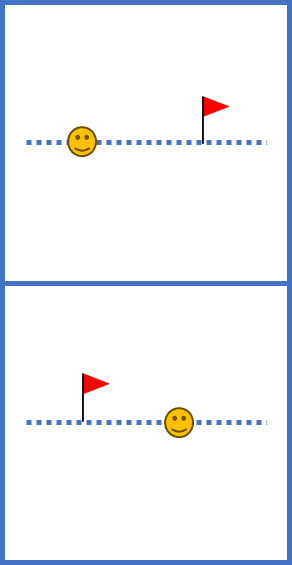} & \includegraphics[height=3cm]{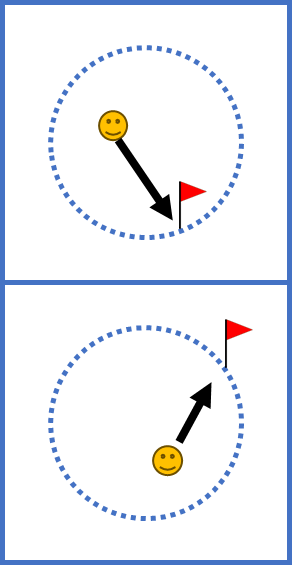} &  \includegraphics[height=3cm]{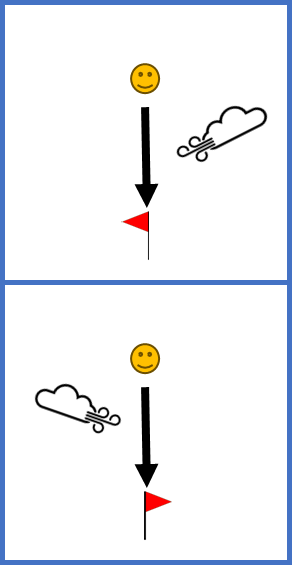} & \includegraphics[height=3cm]{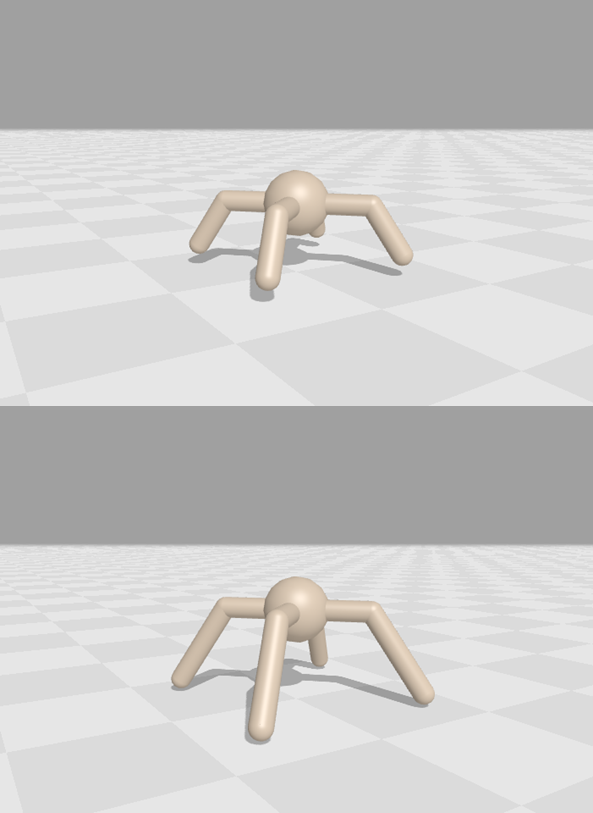} & \includegraphics[height=3cm]{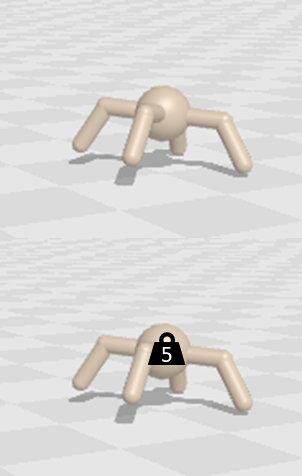} &\includegraphics[trim=2.5cm 0cm 1.5cm 1.5cm, clip, height=3cm]{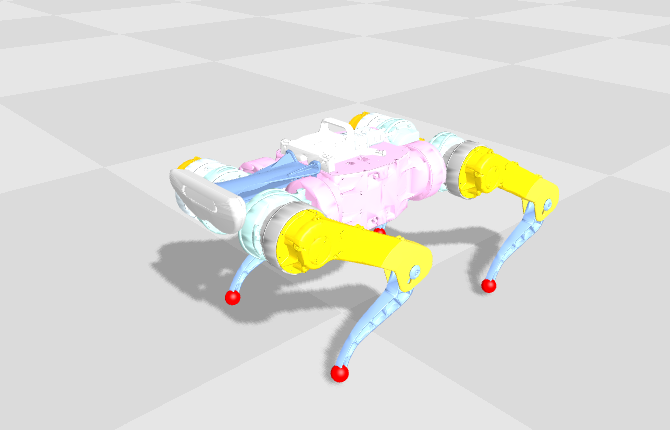}
    \end{tabularx}
    \centering
    \caption{Illustrations of our evaluation environments. From left to right: \textit{1D-Goal}, \textit{2D-Goal}, \textit{2D-Wind}, \textit{Ant-Leg}, \textit{Ant-Weight}, \textit{Barkour-Weight}. In \textit{1D-Goal} and \textit{2D-Goal} the goal location and thus the reward function depends on the HiP $z$. In the remaining tasks the transition function changes. }
    \label{fig:taskvis}
\end{figure}

To validate our approach, we compare its performance with multiple baselines:
1) \textit{Blind}, where we do not add any information to the states in the offline dataset, $\hat{s}=(s)$.
2) \textit{Oracle}, where we augment the dataset with the ground-truth HiP $\hat{s}=(s,z)$. Note that this value is not accessible in practice.
3) \textit{BOReL}, a method that infers the HiP during the episode using a VAE \citep{dorfman_offline_2021_borel}. As we are not able to use reward relabeling or policy replaying for the complete BOReL, we denote it with \textit{BOReL--}.
4) \textit{ContraBAR--}, ContraBAR \citep{choshen_contrabar_2023} addresses the same setting as BOReL, but like us uses a CPC-based architecture. We are not able to use reward relabeling or hard negative mining and therefore denote it \textit{ContraBAR--}.
5) \textit{VRNN}, we also evaluate a Dynamic VAE based method, similar to the LILAC \citep{xie_deep_2021} which was proposed to address the online DP-MDP setting. Our implementation deviates from LILAC by using a Variational RNN \citep{chung_recurrent_2016VRNN} as dynamic VAE, as we found it to perform well in preliminary experiments. We also use TD3-BC to train the policy, while LILAC uses SAC.

To generate our dataset, we train a policy using TD3, or PPO for \textit{Barkour}, in the same DP-MDP without access to the HiP.
As the HiP changes after each episode, this is similar to utilizing domain randomization \citep{tobin_domain_2017}, resulting in a robust policy that is not specialized to any HiP but close to optimal on the "marginal" MDP, i.e., an MDP where the transition density and reward functions are marginalized over $z$.
We then collect the dataset by generating rollouts with the same exploration noise as during training.
Our evaluation tasks are based on related work \citep{xie_deep_2021,dorfman_offline_2021_borel} and we begin with three low-dimensional tasks: 
In \textit{1D-Goal} the agent starts in a random position and navigates to one of two goals, while in \textit{2D-Goal} the agent moves to a goal location on a circle, the exact position of which depends on the HiP $z$.
In \textit{2D-Wind} we instead change the transition function, by adding a disturbance to the location depending on the HiP $z$.
As higher-dimensional tasks we consider two variations of the well known Ant \citep{schulman_high-dimensional_2018} task, simulated using BRAX \citep{freeman_brax_2021}.
In \textit{Ant-Weight} we vary the mass of the body of the ant to simulate a varying load, in \textit{Ant-Leg} we change the length of the legs of the ant to simulate dyanmics change due to wear and tear. 
Finally, in \textit{Barkour-Weight} we validate our approach on a realistic simulation of a full robot. 
We utilize the simulation of the Barkour robot provided by \cite{caluwaerts_barkour_2023}, which they have shown to be successful in sim-to-real transfer. 
We therefore consider it to be a good proxy for real world applications.
Note that the observation here includes the last three observations and actions, $s_t=(o_{t-2},a_{t-2},o_{t-1},a_{t-1},o_t)$, such that the weight could be inferred without any additional input.
We illustrate these tasks in Fig. \ref{fig:taskvis}.

We split our evaluation into three parts: 
1) Does the model learn a useful representation of the HiP?
2) Can we accurately infer the next latent during evaluation?
3) Does this allow us to learn a better policy from the offline dataset?

\paragraph{Learned Representation}
\newcommand{\includegraphicsTrimmed}[1]{%
    \includegraphics[trim=1cm 0.6cm 1.5cm 1.1cm, clip, width=.2\linewidth]{#1}%
}
\begin{figure}
    \centering
    \begin{minipage}{0.69\linewidth}
    \resizebox{\linewidth}{!}{%
        \setlength{\tabcolsep}{0pt}
        \renewcommand{\arraystretch}{0.6}
        \tiny
        \begin{tabular}{c@{\hspace{10pt}}cccc}
            \rotatebox{90}{\phantom{12}2D-Goal} & \includegraphicsTrimmed{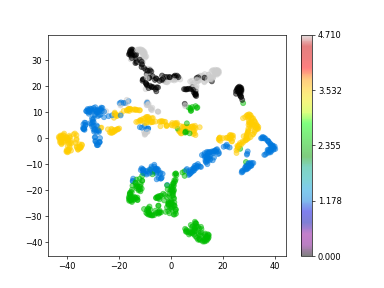} & \includegraphicsTrimmed{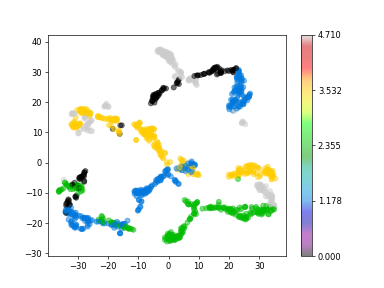} & \includegraphicsTrimmed{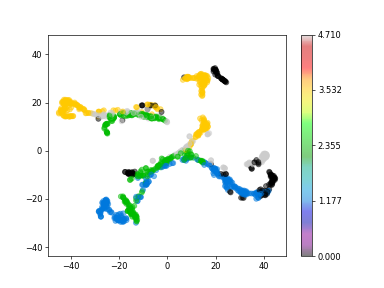} & \includegraphicsTrimmed{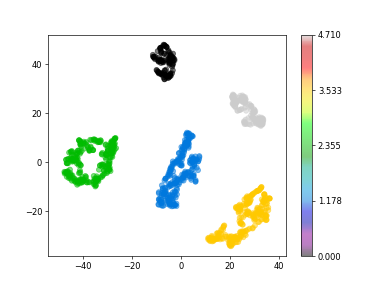}\\
            \rotatebox{90}{Ant-Weight} & \includegraphicsTrimmed{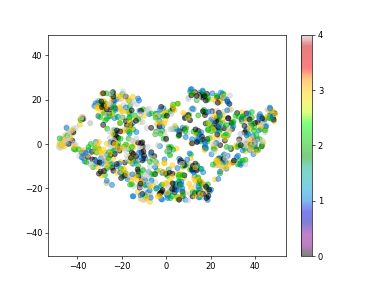} & \includegraphicsTrimmed{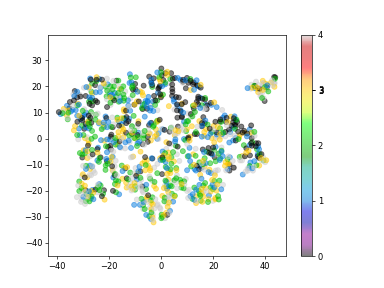} & \includegraphicsTrimmed{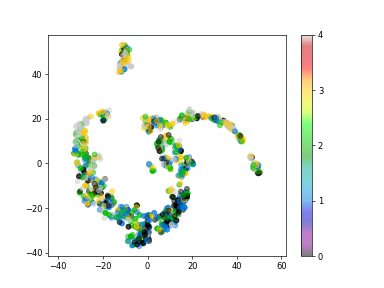} & \includegraphicsTrimmed{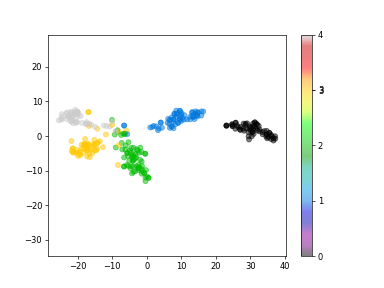}\\
             & BOReL-- & ContraBAR-- & VRNN & COSPA (ours)\\
        \end{tabular}
    }
    \end{minipage}
    \begin{minipage}{0.3\linewidth}
        \begin{tikzpicture}
\definecolor{lightgray204}{RGB}{204,204,204}
\definecolor{darkgray176}{RGB}{176,176,176}

\begin{axis}[
width=\linewidth,
height=4.5cm,
ybar,
tick align=outside,
bar width=0.35,
tick pos=left,
enlargelimits=0.15,
legend style={nodes={scale=0.75, transform shape}},
legend cell align={left},
legend style={
  fill opacity=0.8,
  draw opacity=1,
  text opacity=1,
  at={(0.0,1.05)},
  anchor=north west,
  draw=lightgray204,
},
ylabel={Linear Probe Acc.},
xtick=data,
ymajorgrids,
x grid style={darkgray176},
xtick style={color=black},
xticklabels={ContraBAR,LILAC,BOReL,CPC},
y grid style={darkgray176},
x tick label style={scale=0.7,rotate=45,anchor=east},
y tick label style={scale=0.7},
y label style={scale=0.9},
xticklabels={BOReL--, ContraBAR--, VRNN, COSPA (ours)},
]
\addplot[blue,fill=cpccolor,error bars/.cd,y dir=both,y explicit,]
coordinates{
   (1,0.756) +-(0.0261,0.0261)
   (2,0.713) +-(0.055,0.055)
   (3,0.6662) +-(0.02322,0.02322)
   (4,0.997715736040609) +-(0.0021,0.0021)
};
\addplot[red,fill=vrnncolor,error bars/.cd,y dir=both,y explicit,]
coordinates {
  (1,0.28) +-(0.0195,0.0195)
  (2,0.2802) +- (0.0252,0.0252)
  (3,0.2125) +- (0.0175,0.0175)
  (4,0.934825912918587) +-(0.008891958442132,0.008891958442132)
};
\legend{2D-Goal,Ant-Weight}
\end{axis}
\end{tikzpicture}
    \end{minipage}
    \caption{Comparison of the learned representations. The left side shows T-SNE visualizations, the right side shows the mean test accuracy with 95\% CIs across 20 trials of linear probes trained to predict the ground-truth HiPs. Each dot is the embedding of a trajectory, the HiP of which is represented by the color. For BOReL and VRNN the mean $\mu$ of the posterior is visualized.}
    \label{fig:representationeval}
\end{figure}
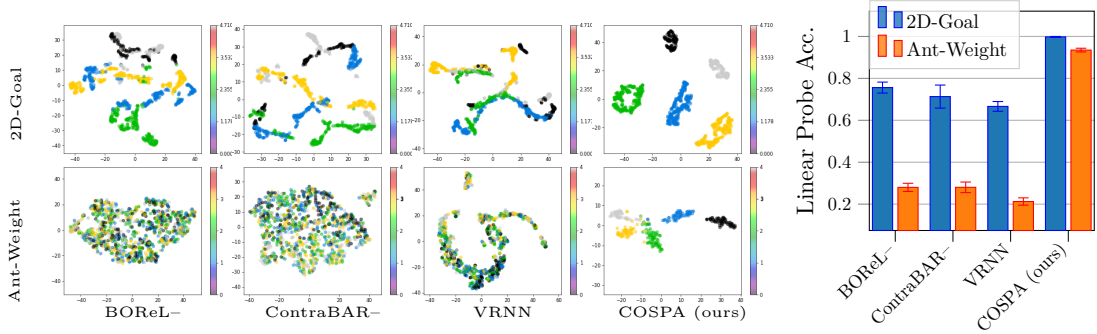
To evaluate our learned representation of the HiP, we follow common practices from the representation learning community \citep{nozawa_empirical_2022}. We quantitatively evaluate the learned representation using linear probes and qualitatively evaluate it using T-SNE \citep{maaten_tsne_2008} to visualize the learned representation.
As we show in Fig. \ref{fig:representationeval}, our method learns a representation which captures the underlying structure well, clustering tasks by HiP.
The baselines perform reasonably well in the 2D-Goal task.
On the more difficult Ant-Weight task, our method performs significantly better. While we can see some structure in the representations learned by the ContraBAR and BOReL baselines, they are not able to learn a useful representation.

\textbf{Next Latent Prediction}
\begin{figure}
    \centering
    \includegraphics[trim=0.4cm 0.3cm 0.5cm 0.5cm, clip, width=0.2\linewidth]{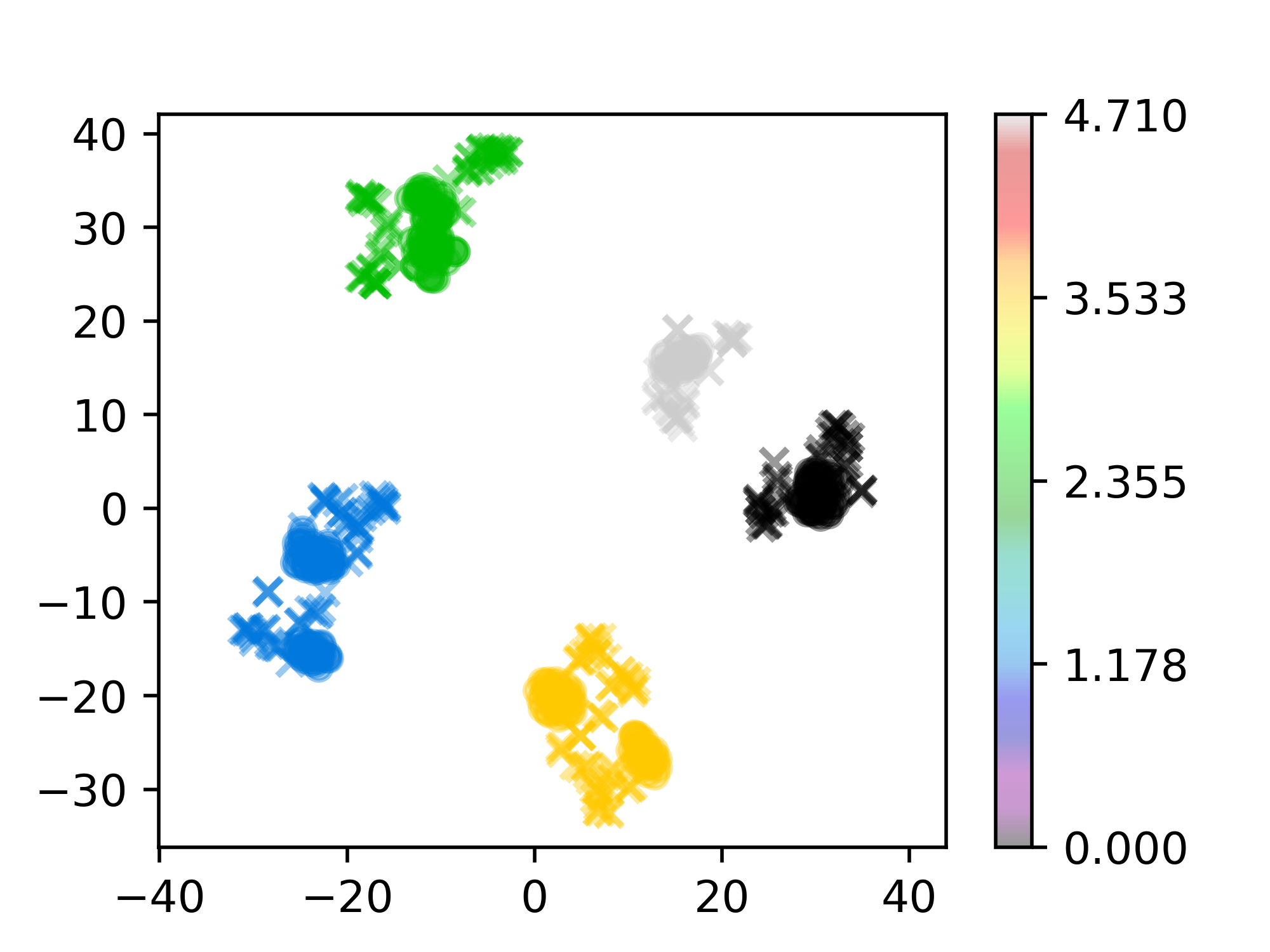}
    \includegraphics[trim=0.4cm 0.3cm 0.5cm 0.5cm, clip, width=0.2\linewidth]{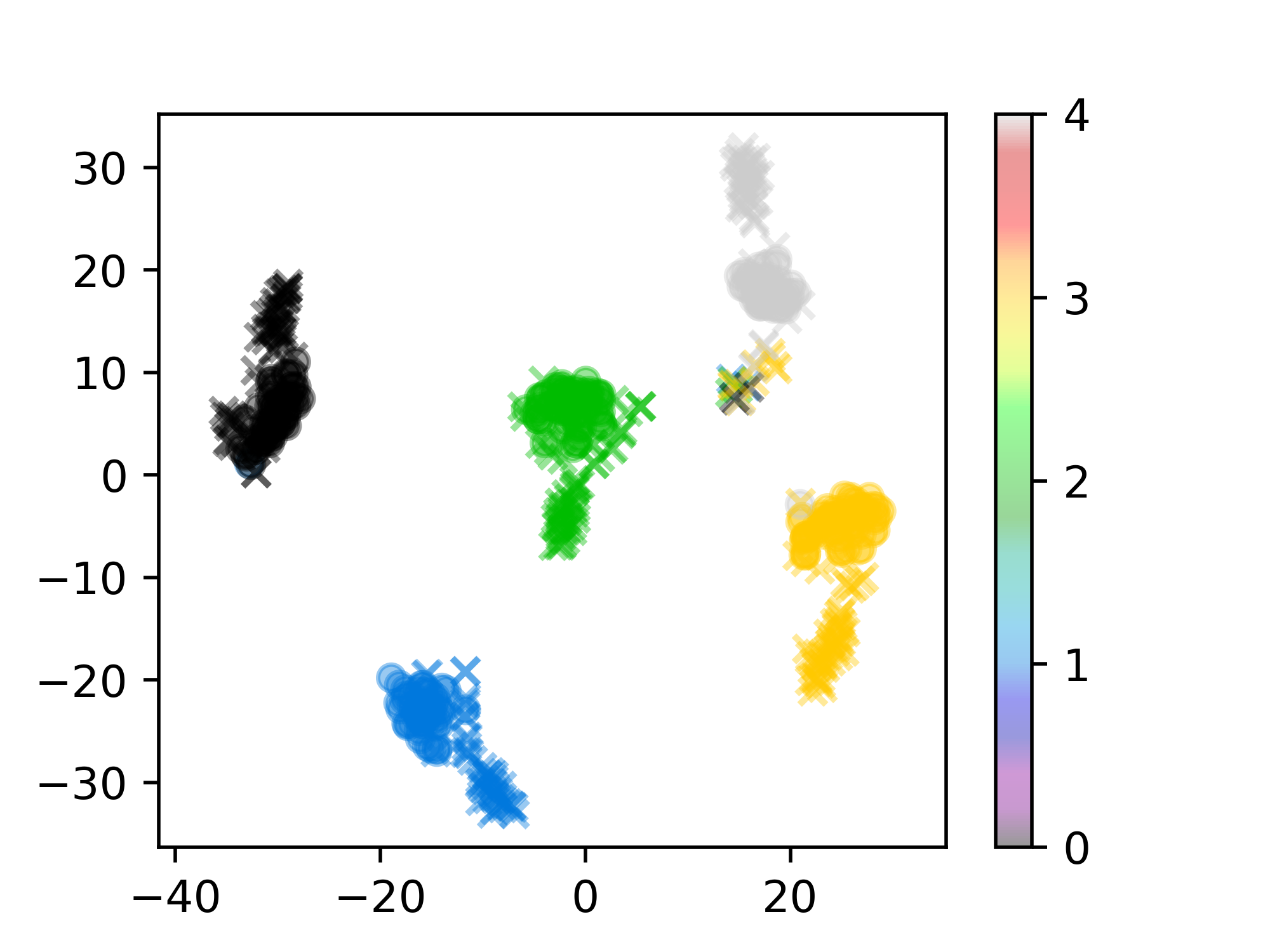}
    \caption{T-SNE visualization of the inferred latents $\tilde{z}$ as crosses ($\times$), and predicted latents $\bar{z}_i=f_\mathrm{pred}(\tilde{z}_{i-N_c},\dots,\tilde{z}_{i-1})$ as circles ($\circ$).}
    \label{fig:predvis}
\end{figure}
As we need to predict the latent $\tilde{z}_i$ from the given context during deployment, we also evaluate the prediction.
We visualize the similarity between predicted latent $\bar{z}_i = f_{pred}(\tilde{z}_{i-N_c:i-1})$ and inferred latents $\tilde{z}_i = g_{\mathrm{enc}}(\tau_i)$ by embedding them in a shared T-SNE visualization.
The results are shown in Fig. \ref{fig:predvis}, and we can see a good correspondence between the inferred and predicted latents.
The predicted latents are more concentrated than the inferred latents, which can be explained by the denoising properties of the regression loss.

\paragraph{Offline RL}
\begin{figure}
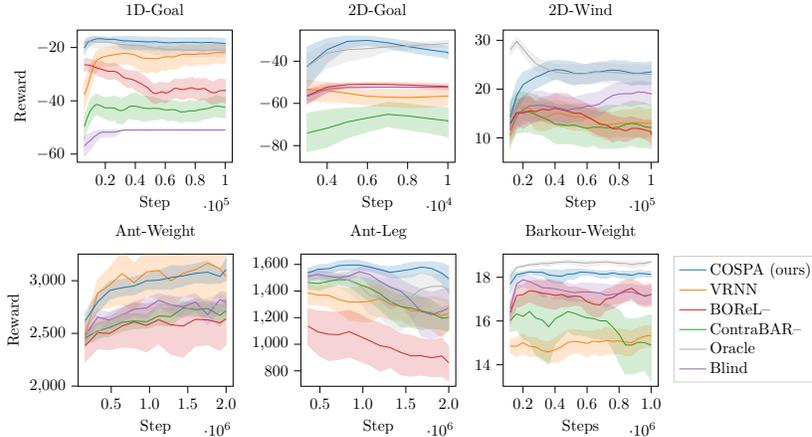

    \centering
    \resizebox{0.75\linewidth}{!}{
    \newlength{\plotwidth}
    \newlength{\plotheight}
    \setlength{\plotwidth}{5cm}
    \setlength{\plotheight}{4.5cm}
    \setlength{\tabcolsep}{0pt}
    \renewcommand{\arraystretch}{0.8}
    \begin{tabular}{cccc}
         \input{pgf/xlrevolvev3smooth} & \begin{tikzpicture}

\definecolor{crimson2143940}{RGB}{214,39,40}
\definecolor{darkgray176}{RGB}{176,176,176}
\definecolor{darkorange25512714}{RGB}{255,127,14}
\definecolor{forestgreen4416044}{RGB}{44,160,44}
\definecolor{lightgray204}{RGB}{204,204,204}
\definecolor{mediumpurple148103189}{RGB}{148,103,189}
\definecolor{sienna1408675}{RGB}{140,86,75}
\definecolor{steelblue31119180}{RGB}{31,119,180}

\begin{axis}[
legend cell align={left},
legend style={
  fill opacity=0.8,
  draw opacity=1,
  text opacity=1,
  at={(0.03,0.03)},
  anchor=south west,
  draw=lightgray204
},
tick align=outside,
tick pos=left,
x grid style={darkgray176},
xlabel={Step},
xmin=2649.8, xmax=10354.2,
xtick style={color=black},
y grid style={darkgray176},
ymin=-87.8951174995899, ymax=-25.2764936139584,
ytick style={color=black},
width=\plotwidth,
height=\plotheight,
title={2D-Goal}
]
\path [draw=contrabarcolor, fill=contrabarcolor, opacity=0.2]
(axis cs:3000,-64.6864071210225)
--(axis cs:3000,-82.863393629392)
--(axis cs:4000,-80.7150599940618)
--(axis cs:5000,-76.3588488483429)
--(axis cs:6000,-73.4407383934657)
--(axis cs:7000,-71.0437562100093)
--(axis cs:8000,-72.4762719678879)
--(axis cs:9000,-73.9037251869837)
--(axis cs:10000,-75.7926314179103)
--(axis cs:10004,-76.0849919907252)
--(axis cs:10004,-62.1436709909439)
--(axis cs:10004,-62.1436709909439)
--(axis cs:10000,-62.1042541980743)
--(axis cs:9000,-61.4707206471761)
--(axis cs:8000,-60.1959283240636)
--(axis cs:7000,-59.4835954697927)
--(axis cs:6000,-61.7600226910909)
--(axis cs:5000,-62.2205812978744)
--(axis cs:4000,-63.084763563474)
--(axis cs:3000,-64.6864071210225)
--cycle;

\path [draw=vrnncolor, fill=vrnncolor, opacity=0.2]
(axis cs:3000,-48.9872792069117)
--(axis cs:3000,-58.8832839075724)
--(axis cs:4000,-59.2290633916855)
--(axis cs:5000,-59.6991802597046)
--(axis cs:6000,-60.9881908957164)
--(axis cs:7000,-61.3862439441681)
--(axis cs:8000,-61.995312110583)
--(axis cs:9000,-61.9384740988413)
--(axis cs:10000,-60.8737790139516)
--(axis cs:10004,-60.4656967207591)
--(axis cs:10004,-52.73606108284)
--(axis cs:10004,-52.73606108284)
--(axis cs:10000,-52.7928800217311)
--(axis cs:9000,-53.2170086510976)
--(axis cs:8000,-53.5174498081207)
--(axis cs:7000,-53.3990842437744)
--(axis cs:6000,-53.1061365826925)
--(axis cs:5000,-51.5661082601547)
--(axis cs:4000,-50.1600987752279)
--(axis cs:3000,-48.9872792069117)
--cycle;

\path [draw=borelcolor, fill=borelcolor, opacity=0.2]
(axis cs:3000,-53.3476892868678)
--(axis cs:3000,-60.1789906597137)
--(axis cs:4000,-54.5049916315079)
--(axis cs:5000,-51.942091887792)
--(axis cs:6000,-51.8366314522425)
--(axis cs:7000,-52.0267792145411)
--(axis cs:8000,-52.486649518013)
--(axis cs:9000,-53.0920339934031)
--(axis cs:10000,-53.4456075191498)
--(axis cs:10004,-53.5062761726379)
--(axis cs:10004,-50.9072642335892)
--(axis cs:10004,-50.9072642335892)
--(axis cs:10000,-50.7561215718587)
--(axis cs:9000,-50.5373203134537)
--(axis cs:8000,-50.285225666364)
--(axis cs:7000,-50.0637038119634)
--(axis cs:6000,-50.0641947412491)
--(axis cs:5000,-50.2631655534109)
--(axis cs:4000,-50.7289846436183)
--(axis cs:3000,-53.3476892868678)
--cycle;

\path [draw=cpccolor, fill=cpccolor, opacity=0.2]
(axis cs:3000,-33.1277202161153)
--(axis cs:3000,-53.1938424412409)
--(axis cs:4000,-40.7987979078293)
--(axis cs:5000,-34.0004809761047)
--(axis cs:6000,-32.4907890407244)
--(axis cs:7000,-33.6468732984861)
--(axis cs:8000,-35.6643232162793)
--(axis cs:9000,-37.2265519913038)
--(axis cs:10000,-38.6788650512695)
--(axis cs:10004,-39.1849210476875)
--(axis cs:10004,-33.6772412980398)
--(axis cs:10004,-33.6772412980398)
--(axis cs:10000,-33.0133060399691)
--(axis cs:9000,-31.8869114804268)
--(axis cs:8000,-30.5198576148351)
--(axis cs:7000,-29.2151746670405)
--(axis cs:6000,-28.1735078414281)
--(axis cs:5000,-28.1227946996689)
--(axis cs:4000,-29.2130100695292)
--(axis cs:3000,-33.1277202161153)
--cycle;

\path [draw=oraclecolor, fill=oraclecolor, opacity=0.2]
(axis cs:3000,-36.6386189254125)
--(axis cs:3000,-50.523573041757)
--(axis cs:4000,-43.2266855192184)
--(axis cs:5000,-40.0706182583173)
--(axis cs:6000,-38.2783456126849)
--(axis cs:7000,-35.2726221776009)
--(axis cs:8000,-34.7261651007334)
--(axis cs:9000,-33.552276515166)
--(axis cs:10000,-33.1931439177195)
--(axis cs:10004,-32.9380866568883)
--(axis cs:10004,-30.0177313450178)
--(axis cs:10004,-30.0177313450178)
--(axis cs:10000,-30.2303662467003)
--(axis cs:9000,-30.8500916306178)
--(axis cs:8000,-31.4517030493418)
--(axis cs:7000,-31.416455997626)
--(axis cs:6000,-31.189380458196)
--(axis cs:5000,-30.817677248319)
--(axis cs:4000,-31.5387134083112)
--(axis cs:3000,-36.6386189254125)
--cycle;

\path [draw=blindcolor, fill=blindcolor, opacity=0.2]
(axis cs:3000,-53.5781834538778)
--(axis cs:3000,-60.4313522847493)
--(axis cs:4000,-54.9038344923655)
--(axis cs:5000,-52.7087976694107)
--(axis cs:6000,-52.4821600341797)
--(axis cs:7000,-52.403957880338)
--(axis cs:8000,-52.4153718058268)
--(axis cs:9000,-52.4178315989176)
--(axis cs:10000,-52.424218228658)
--(axis cs:10004,-52.4637828699748)
--(axis cs:10004,-52.2429061155319)
--(axis cs:10004,-52.2429061155319)
--(axis cs:10000,-52.2591655127207)
--(axis cs:9000,-52.2411036109924)
--(axis cs:8000,-52.2639742883046)
--(axis cs:7000,-52.3101943429311)
--(axis cs:6000,-52.3062538957596)
--(axis cs:5000,-52.3143425655365)
--(axis cs:4000,-52.2500117254257)
--(axis cs:3000,-53.5781834538778)
--cycle;

\addplot [semithick, contrabarcolor, forget plot]
table {%
3000 -74.0476746877034
4000 -71.7407634099325
5000 -68.796617825826
6000 -66.9311033884684
7000 -65.1898599624634
8000 -65.9398429870606
9000 -67.1403343200684
10000 -68.252298227946
10004 -68.7742893346151
};
\addplot [semithick, vrnncolor, forget plot]
table {%
3000 -53.3888158798218
4000 -53.9948320388794
5000 -55.19087378184
6000 -56.5746611913045
7000 -57.0593191782633
8000 -57.2186336517334
9000 -56.9000752131144
10000 -56.5452110290527
10004 -56.1970365524292
};
\addplot [semithick, borelcolor, forget plot]
table {%
3000 -56.318151028951
4000 -52.2562331517537
5000 -51.1002224604289
6000 -50.9161975224813
7000 -51.0132141749064
8000 -51.3111010869344
9000 -51.6654691696167
10000 -51.9716571172078
10004 -52.0917148717244
};
\addplot [semithick, cpccolor, forget plot]
table {%
3000 -42.4616105079651
4000 -34.4710446039836
5000 -30.6293280601501
6000 -30.1545611699422
7000 -31.2574435869853
8000 -33.0385849952698
9000 -34.5380328496297
10000 -35.8666674931844
10004 -36.3916244443258
};
\addplot [semithick, oraclecolor, forget plot]
table {%
3000 -42.5829848607381
4000 -36.1754050890605
5000 -34.6021381696065
6000 -34.0785880724589
7000 -33.2262658437093
8000 -32.9207061449687
9000 -32.166804377238
10000 -31.639620812734
10004 -31.4775020853678
};
\addplot [semithick, blindcolor, forget plot]
table {%
3000 -56.7332534154256
4000 -53.1654996236165
5000 -52.471365737915
6000 -52.3896104176839
7000 -52.3576424916585
8000 -52.335500907898
9000 -52.3238629659017
10000 -52.3336315155029
10004 -52.337954750061
};
\end{axis}

\end{tikzpicture} & \input{pgf/xywindsmooth} &  \\
         \begin{tikzpicture}

\definecolor{crimson2143940}{RGB}{214,39,40}
\definecolor{darkgray176}{RGB}{176,176,176}
\definecolor{darkorange25512714}{RGB}{255,127,14}
\definecolor{forestgreen4416044}{RGB}{44,160,44}
\definecolor{lightgray204}{RGB}{204,204,204}
\definecolor{mediumpurple148103189}{RGB}{148,103,189}
\definecolor{sienna1408675}{RGB}{140,86,75}
\definecolor{steelblue31119180}{RGB}{31,119,180}

\begin{axis}[
legend cell align={left},
legend style={
  fill opacity=0.8,
  draw opacity=1,
  text opacity=1,
  at={(0.03,0.03)},
  anchor=south west,
  draw=lightgray204
},
tick align=outside,
tick pos=left,
x grid style={darkgray176},
xlabel={Step},
xmin=67999.8, xmax=2092004.2,
xtick style={color=black},
xtick={0,500000,1000000,1500000,2000000,2500000},
xticklabels={0.0,0.5,1.0,1.5,2.0,2.5},
y grid style={darkgray176},
ylabel={Reward},
ymin=1998.12522392426, ymax=3252.21344914093,
ytick style={color=black},
width=\plotwidth,
height=\plotheight,
title={Ant-Weight}
]
\path [draw=vrnncolor, fill=vrnncolor, opacity=0.2]
(axis cs:160000,2603.66826126099)
--(axis cs:160000,2375.41844863892)
--(axis cs:320000,2729.23107727051)
--(axis cs:480000,2745.18233032227)
--(axis cs:640000,2778.78739715576)
--(axis cs:800000,2709.99544403076)
--(axis cs:960000,2860.2690045166)
--(axis cs:1120000,2825.18104812622)
--(axis cs:1280000,2721.5858821106)
--(axis cs:1440000,2857.87882461548)
--(axis cs:1600000,2835.71935791016)
--(axis cs:1760000,2826.21318511963)
--(axis cs:1920000,2835.3434197998)
--(axis cs:2000004,2746.13927935791)
--(axis cs:2000004,3246.64167364502)
--(axis cs:2000004,3246.64167364502)
--(axis cs:1920000,3322.13181274414)
--(axis cs:1760000,3393.36918579102)
--(axis cs:1600000,3296.16197036743)
--(axis cs:1440000,3247.37697692871)
--(axis cs:1280000,3231.3975579834)
--(axis cs:1120000,3290.1414743042)
--(axis cs:960000,3270.79662353516)
--(axis cs:800000,3171.61773834229)
--(axis cs:640000,3271.7611428833)
--(axis cs:480000,3124.96066711426)
--(axis cs:320000,3005.78320892334)
--(axis cs:160000,2603.66826126099)
--cycle;

\path [draw=contrabarcolor, fill=contrabarcolor, opacity=0.2]
(axis cs:160000,2521.41039474487)
--(axis cs:160000,2393.62853973389)
--(axis cs:320000,2455.87194412231)
--(axis cs:480000,2472.41847305298)
--(axis cs:640000,2520.49657867432)
--(axis cs:800000,2545.05851974487)
--(axis cs:960000,2549.61504745483)
--(axis cs:1120000,2582.41699584961)
--(axis cs:1280000,2566.9278150177)
--(axis cs:1440000,2677.14850845337)
--(axis cs:1600000,2663.94730728149)
--(axis cs:1760000,2658.05265319824)
--(axis cs:1920000,2572.01970291138)
--(axis cs:2000004,2644.05438339233)
--(axis cs:2000004,2794.24662298584)
--(axis cs:2000004,2794.24662298584)
--(axis cs:1920000,2759.56178924561)
--(axis cs:1760000,2806.9076121521)
--(axis cs:1600000,2783.59277328491)
--(axis cs:1440000,2803.42889724731)
--(axis cs:1280000,2738.13064704895)
--(axis cs:1120000,2747.51524414062)
--(axis cs:960000,2672.98506057739)
--(axis cs:800000,2686.71459960938)
--(axis cs:640000,2683.68462600708)
--(axis cs:480000,2630.91273452759)
--(axis cs:320000,2594.63934799194)
--(axis cs:160000,2521.41039474487)
--cycle;

\path [draw=borelcolor, fill=borelcolor, opacity=0.2]
(axis cs:160000,2507.21555671692)
--(axis cs:160000,2232.1056215922)
--(axis cs:320000,2352.76209665934)
--(axis cs:480000,2345.35153745015)
--(axis cs:640000,2369.94787305196)
--(axis cs:800000,2364.36135724386)
--(axis cs:960000,2323.26523113251)
--(axis cs:1120000,2401.7424765269)
--(axis cs:1280000,2300.64269065857)
--(axis cs:1440000,2276.14419911702)
--(axis cs:1600000,2380.24744822184)
--(axis cs:1760000,2390.56021664937)
--(axis cs:1920000,2339.1154355367)
--(axis cs:2000004,2392.4430725352)
--(axis cs:2000004,2825.78516494751)
--(axis cs:2000004,2825.78516494751)
--(axis cs:1920000,2794.2852839152)
--(axis cs:1760000,2815.94779434204)
--(axis cs:1600000,2826.19582010905)
--(axis cs:1440000,2775.45471776326)
--(axis cs:1280000,2761.03073209127)
--(axis cs:1120000,2796.12952575684)
--(axis cs:960000,2772.26630859375)
--(axis cs:800000,2790.39259223938)
--(axis cs:640000,2738.75237553914)
--(axis cs:480000,2657.02186470032)
--(axis cs:320000,2648.12196667989)
--(axis cs:160000,2507.21555671692)
--cycle;

\path [draw=cpccolor, fill=cpccolor, opacity=0.2]
(axis cs:160000,2699.46413452148)
--(axis cs:160000,2539.91445526123)
--(axis cs:320000,2699.45495666504)
--(axis cs:480000,2824.57342346191)
--(axis cs:640000,2858.90648345947)
--(axis cs:800000,2867.10789581299)
--(axis cs:960000,2917.44482910156)
--(axis cs:1120000,2918.14248199463)
--(axis cs:1280000,2919.41329162598)
--(axis cs:1440000,2950.40025695801)
--(axis cs:1600000,2980.91666320801)
--(axis cs:1760000,2971.77049591064)
--(axis cs:1920000,2975.64757080078)
--(axis cs:2000004,3002.98218847656)
--(axis cs:2000004,3204.72557452393)
--(axis cs:2000004,3204.72557452393)
--(axis cs:1920000,3106.70338592529)
--(axis cs:1760000,3188.3323046875)
--(axis cs:1600000,3157.84879455566)
--(axis cs:1440000,3137.98)
--(axis cs:1280000,3143.3153024292)
--(axis cs:1120000,3094.74629760742)
--(axis cs:960000,3067.27424682617)
--(axis cs:800000,3036.46829376221)
--(axis cs:640000,3014.14807067871)
--(axis cs:480000,2993.26769683838)
--(axis cs:320000,2882.46491729736)
--(axis cs:160000,2699.46413452148)
--cycle;

\path [draw=blindcolor, fill=blindcolor, opacity=0.2]
(axis cs:160000,2546.03802932213)
--(axis cs:160000,2408.85516441608)
--(axis cs:320000,2588.17586733062)
--(axis cs:480000,2548.72084224306)
--(axis cs:640000,2610.09121914568)
--(axis cs:800000,2666.25907256028)
--(axis cs:960000,2643.90060551084)
--(axis cs:1120000,2748.46951357085)
--(axis cs:1280000,2694.3275396939)
--(axis cs:1440000,2650.83425671808)
--(axis cs:1600000,2715.13098965349)
--(axis cs:1760000,2582.36017845417)
--(axis cs:1920000,2717.2925096183)
--(axis cs:2000004,2691.65029120083)
--(axis cs:2000004,2884.59082837335)
--(axis cs:2000004,2884.59082837335)
--(axis cs:1920000,2896.21798537682)
--(axis cs:1760000,2769.25290969322)
--(axis cs:1600000,2878.44620677027)
--(axis cs:1440000,2845.4973462335)
--(axis cs:1280000,2845.86027453192)
--(axis cs:1120000,2876.65375239931)
--(axis cs:960000,2836.80037484005)
--(axis cs:800000,2802.17981672616)
--(axis cs:640000,2739.52326807482)
--(axis cs:480000,2698.09276522932)
--(axis cs:320000,2730.30161132813)
--(axis cs:160000,2546.03802932213)
--cycle;

\path [draw=oraclecolor, fill=oraclecolor, opacity=0.2]
(axis cs:160000,2610.89094238281)
--(axis cs:160000,2417.45640194542)
--(axis cs:320000,2426.8812969007)
--(axis cs:480000,2539.25352044356)
--(axis cs:640000,2572.03105147512)
--(axis cs:800000,2592.07520205849)
--(axis cs:960000,2554.38668983861)
--(axis cs:1120000,2689.06520482114)
--(axis cs:1280000,2647.06425331517)
--(axis cs:1440000,2652.06622924805)
--(axis cs:1600000,2654.56117682206)
--(axis cs:1760000,2654.04148655942)
--(axis cs:1920000,2678.88529020611)
--(axis cs:2000004,2652.68723703485)
--(axis cs:2000004,2840.36537218596)
--(axis cs:2000004,2840.36537218596)
--(axis cs:1920000,2875.94134617856)
--(axis cs:1760000,2828.01855083265)
--(axis cs:1600000,2832.69596300627)
--(axis cs:1440000,2792.69265008224)
--(axis cs:1280000,2866.72845812346)
--(axis cs:1120000,2840.50507459139)
--(axis cs:960000,2785.09569766396)
--(axis cs:800000,2776.71515920539)
--(axis cs:640000,2767.27782592773)
--(axis cs:480000,2701.63923950195)
--(axis cs:320000,2593.2537253932)
--(axis cs:160000,2610.89094238281)
--cycle;

\addplot [semithick, vrnncolor, forget plot]
table {%
160000 2491.53274536133
320000 2877.7630859375
480000 2959.09729614258
640000 3067.11131591797
800000 2981.89590454102
960000 3083.66271972656
1120000 3097.5510925293
1280000 2997.38888549805
1440000 3071.34276733398
1600000 3104.9037902832
1760000 3165.7248046875
1920000 3114.66235351562
2000004 3035.36538024902
};
\addplot [semithick, contrabarcolor, forget plot]
table {%
160000 2457.33128051758
320000 2521.63020019531
480000 2557.06201782227
640000 2604.00229492188
800000 2617.46679077148
960000 2610.03892211914
1120000 2665.36030273438
1280000 2660.10897521973
1440000 2741.54544677734
1600000 2724.99748535156
1760000 2735.92599487305
1920000 2670.54326477051
2000004 2715.13345031738
};
\addplot [semithick, borelcolor, forget plot]
table {%
160000 2382.64363098145
320000 2505.24137369792
480000 2499.98305765788
640000 2569.03907775879
800000 2609.10897827148
960000 2575.92603810628
1120000 2629.96344502767
1280000 2581.47950744629
1440000 2570.49810155233
1600000 2633.42850240072
1760000 2627.96276601156
1920000 2600.34620412191
2000004 2637.45844014486
};
\addplot [semithick, cpccolor, forget plot]
table {%
160000 2619.53227539062
320000 2797.32766113281
480000 2908.16253662109
640000 2933.13645019531
800000 2950.6064453125
960000 2998.36857910156
1120000 3005.72196044922
1280000 3029.87503662109
1440000 3046.08262939453
1600000 3069.59484863281
1760000 3080.77446289062
1920000 3039.67513427734
2000004 3104.39741943359
};
\addplot [semithick, blindcolor, forget plot]
table {%
160000 2479.24232219828
320000 2654.46442281789
480000 2625.91562836746
640000 2674.95900121228
800000 2729.66835600754
960000 2740.77207367996
1120000 2814.508847993
1280000 2770.31012594289
1440000 2751.9923474542
1600000 2798.3876953125
1760000 2676.89738937904
1920000 2820.03398184941
2000004 2797.27484762258
};
\addplot [semithick, oraclecolor, forget plot]
table {%
160000 2510.85226922286
320000 2506.37449886924
480000 2618.85584138569
640000 2677.6655337685
800000 2685.83844315378
960000 2669.47314453125
1120000 2767.50809518914
1280000 2764.93400493421
1440000 2720.89711400082
1600000 2751.97938939145
1760000 2743.92885228207
1920000 2774.14392732319
2000004 2750.1855417352
};
\end{axis}

\end{tikzpicture} & \input{pgf/antleglensmooth} & \multicolumn{2}{c}{\input{pgf/bark-weight-noangrew-smooth}} 
    \end{tabular}
    }
    \caption{Results of our proposed method on the evaluation environments. Shown are the mean evaluation rewards for 20 seeds per experiment, the shaded areas show 95\% confidence intervals.}
    \label{fig:allresults}
\end{figure}
\begin{table}[]
    \centering
    \resizebox{\linewidth}{!}{
    \begin{tabular}{c||c c|c c c c}
                & Blind           &   Oracle        & VRNN-             & BOReL             &   ContraBAR-       & COSPA (ours)          \\
       \hline
       1D-Goal & $-50.95\pm0.02$ & $-20.79\pm2.40$  & $\mathbf{-21.86\pm3.96}$   & $-36.41\pm4.32$   & $-42.43\pm3.85$ & $\mathbf{-18.59\pm1.83}$  \\
       2D-Goal & $-52.34 \pm0.10$ & $-31.48\pm1.46$ & $-56.20\pm3.47$   & $-52.09\pm1.18$   & $-68.77\pm6.63$ & $\mathbf{-36.39\pm2.72}$  \\
       2D-Wind & $18.88\pm2.03$ & $22.81\pm2.01$    & $13.03\pm2.75$    & $10.63\pm2.17$    & $12.04\pm4.45$  & $\mathbf{23.63\pm2.38}$  \\
       Ant-Weight & $2797\pm87$ & $2750\pm98$       & $\mathbf{3035\pm211}$& $2637\pm188$   & $2715\pm79$     & $\mathbf{3104\pm100}$  \\
       Ant-Leg & $\mathbf{1273\pm142}$ & $1407\pm92$         & $1231\pm98$       & $862\pm121$       & $1201\pm92$     & $\mathbf{1493\pm97}$  \\
       Barkour-Weight& $17.19\pm0.40$ & $18.71\pm0.05$ & $15.34\pm0.46$ & $17.22\pm0.50$ & $14.90\pm1.38$ & $\mathbf{18.13\pm0.14}$  \\
    \end{tabular}
    }
    \caption{Evaluation reward at the end of the training. Mean and 95\% CI across 20 trials. The best performing method and overlapping CIs without access to privileged information are printed \textbf{bold}.}
    \label{tab:my_label}
\end{table}
Having shown that our method is able to identify a useful latent and predict it during inference, we now evaluate the performance of policies trained with the augmented state $\hat{s}=(s,\tilde{z})$.
The results are shown in Fig. \ref{fig:allresults}.
Overall, our proposed method performs well, matching or exceeding the Oracle performance in most tasks.
Interestingly, our proposed method and VRNN perform better than the Oracle in the \textit{Ant-Weight} task. 
One explanation for this is that the learned representation is better able to represent similarity between different HiPs than the simple ground-truth weight value.
While the baselines perform well in the simple \textit{1D-Goal} task, they generally do not perform well in the more difficult settings, sometimes performing worse than the \textit{blind} baseline.
As noted above, in the \textit{Barkour-Weight} task the state includes the previous two observations and actions, in principle allowing even the \textit{blind} baseline to perform optimally.
However, we find that in practice our method still performs significantly better with the inclusion of the inferred latent $\tilde{z}$.
\paragraph{ContraBAR and BOReL}
The relatively poor performance of the ContraBAR-- and BOReL-- baselines in some of our experiments might be surprising, we therefore highlight some differences between their settings and ours to explain the difference in performance.
Compared to BOReL and (offline) ContraBAR, one signficant difference is the nature of the datasets and environments.
While BOReL and offline ContraBAR were designed for datasets generated by task-specific policies for each different HiP, our datasets are generated by a HiP-agnostic policy, making the task inference more challenging.
Further, the majority of the experiments by \cite{dorfman_offline_2021_borel} and \cite{choshen_contrabar_2023} focus on settings with varying reward functions, while our setting focuses on settings with varying transition functions.
Finally, as outlined above, the techniques of \textit{reward relabeling}, \textit{policy replaying} and \textit{hard negative mining} are not applicable in our setting and are shown to be important in the ablations in \cite{dorfman_offline_2021_borel,choshen_contrabar_2023}.

\subsection{Further Experiments}
\begin{figure}
    \begin{minipage}{0.5\linewidth}
        \input{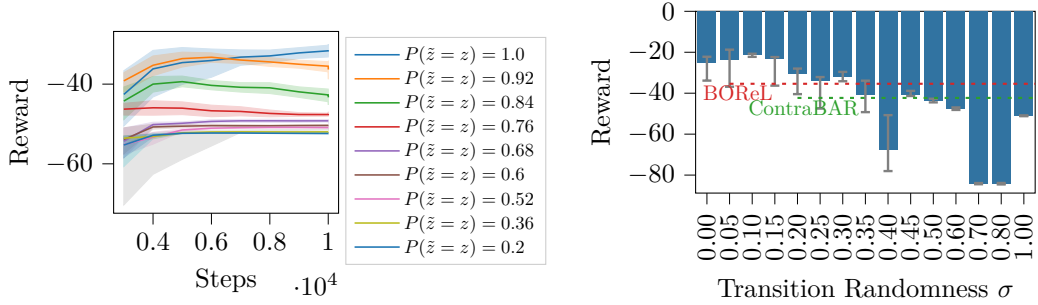}
    \end{minipage}
    \begin{minipage}{0.4\linewidth}
        \begin{tikzpicture}

\definecolor{darkgray176}{RGB}{176,176,176}
\definecolor{steelblue49115161}{RGB}{49,115,161}

\begin{axis}[
tick align=outside,
tick pos=left,
unbounded coords=jump,
x grid style={darkgray176},
xlabel={Transition Randomness $\sigma$},
xmin=-0.5, xmax=14.5,
xtick style={color=black},
xtick={0,1,2,3,4,5,6,7,8,9,10,11,12,13,14},
xticklabels={0.00,0.05,0.10,0.15,0.20,0.25,0.30,0.35,0.40,0.45,0.50,0.60,0.70,0.80,1.00},
y grid style={darkgray176},
ylabel={Reward},
ymin=-88.7877731323242, ymax=0,
ytick style={color=black},
height=4cm,
width=\linewidth,
x tick label style={rotate=90,anchor=east}
]
\draw[draw=none,fill=steelblue49115161] (axis cs:-0.4,0) rectangle (axis cs:0.4,-25.3983936309814);
\draw[draw=none,fill=steelblue49115161] (axis cs:0.6,0) rectangle (axis cs:1.4,-23.868579864502);
\draw[draw=none,fill=steelblue49115161] (axis cs:1.6,0) rectangle (axis cs:2.4,-21.2328243255615);
\draw[draw=none,fill=steelblue49115161] (axis cs:2.6,0) rectangle (axis cs:3.4,-23.4088020324707);
\draw[draw=none,fill=steelblue49115161] (axis cs:3.6,0) rectangle (axis cs:4.4,-30.69091796875);
\draw[draw=none,fill=steelblue49115161] (axis cs:4.6,0) rectangle (axis cs:5.4,-33.9759330749512);
\draw[draw=none,fill=steelblue49115161] (axis cs:5.6,0) rectangle (axis cs:6.4,-32.0842399597168);
\draw[draw=none,fill=steelblue49115161] (axis cs:6.6,0) rectangle (axis cs:7.4,-40.7983283996582);
\draw[draw=none,fill=steelblue49115161] (axis cs:7.6,0) rectangle (axis cs:8.4,-67.8122711181641);
\draw[draw=none,fill=steelblue49115161] (axis cs:8.6,0) rectangle (axis cs:9.4,-40.990062713623);
\draw[draw=none,fill=steelblue49115161] (axis cs:9.6,0) rectangle (axis cs:10.4,-43.9206581115723);
\draw[draw=none,fill=steelblue49115161] (axis cs:10.6,0) rectangle (axis cs:11.4,-47.5625190734863);
\draw[draw=none,fill=steelblue49115161] (axis cs:11.6,0) rectangle (axis cs:12.4,-84.4449920654297);
\draw[draw=none,fill=steelblue49115161] (axis cs:12.6,0) rectangle (axis cs:13.4,-84.4204940795898);
\draw[draw=none,fill=steelblue49115161] (axis cs:13.6,0) rectangle (axis cs:14.4,-51.1234855651855);
\addplot [line width=0.9pt, gray]
table {%
-0.2 -33.8526191711426
0.2 -33.8526191711426
nan nan
0 -33.8526191711426
0 -22.1984596252441
nan nan
-0.2 -22.1984596252441
0.2 -22.1984596252441
};
\addplot [line width=0.9pt, gray]
table {%
0.8 -36.8494071960449
1.2 -36.8494071960449
nan nan
1 -36.8494071960449
1 -18.7319316864014
nan nan
0.8 -18.7319316864014
1.2 -18.7319316864014
};
\addplot [line width=0.9pt, gray]
table {%
1.8 -22.2527256011963
2.2 -22.2527256011963
nan nan
2 -22.2527256011963
2 -20.723180770874
nan nan
1.8 -20.723180770874
2.2 -20.723180770874
};
\addplot [line width=0.9pt, gray]
table {%
2.8 -36.3147163391113
3.2 -36.3147163391113
nan nan
3 -36.3147163391113
3 -22.3429679870605
nan nan
2.8 -22.3429679870605
3.2 -22.3429679870605
};
\addplot [line width=0.9pt, gray]
table {%
3.8 -40.4575996398926
4.2 -40.4575996398926
nan nan
4 -40.4575996398926
4 -27.973596572876
nan nan
3.8 -27.973596572876
4.2 -27.973596572876
};
\addplot [line width=0.9pt, gray]
table {%
4.8 -47.5000381469727
5.2 -47.5000381469727
nan nan
5 -47.5000381469727
5 -32.1124877929688
nan nan
4.8 -32.1124877929688
5.2 -32.1124877929688
};
\addplot [line width=0.9pt, gray]
table {%
5.8 -34.1568984985352
6.2 -34.1568984985352
nan nan
6 -34.1568984985352
6 -29.5916938781738
nan nan
5.8 -29.5916938781738
6.2 -29.5916938781738
};
\addplot [line width=0.9pt, gray]
table {%
6.8 -49.2516593933105
7.2 -49.2516593933105
nan nan
7 -49.2516593933105
7 -33.8861083984375
nan nan
6.8 -33.8861083984375
7.2 -33.8861083984375
};
\addplot [line width=0.9pt, gray]
table {%
7.8 -78.0406265258789
8.2 -78.0406265258789
nan nan
8 -78.0406265258789
8 -50.7198944091797
nan nan
7.8 -50.7198944091797
8.2 -50.7198944091797
};
\addplot [line width=0.9pt, gray]
table {%
8.8 -41.3752670288086
9.2 -41.3752670288086
nan nan
9 -41.3752670288086
9 -38.8596000671387
nan nan
8.8 -38.8596000671387
9.2 -38.8596000671387
};
\addplot [line width=0.9pt, gray]
table {%
9.8 -44.4802093505859
10.2 -44.4802093505859
nan nan
10 -44.4802093505859
10 -42.7197723388672
nan nan
9.8 -42.7197723388672
10.2 -42.7197723388672
};
\addplot [line width=0.9pt, gray]
table {%
10.8 -48.1709060668945
11.2 -48.1709060668945
nan nan
11 -48.1709060668945
11 -46.947322845459
nan nan
10.8 -46.947322845459
11.2 -46.947322845459
};
\addplot [line width=0.9pt, gray]
table {%
11.8 -84.4728393554688
12.2 -84.4728393554688
nan nan
12 -84.4728393554688
12 -83.8881988525391
nan nan
11.8 -83.8881988525391
12.2 -83.8881988525391
};
\addplot [line width=0.9pt, gray]
table {%
12.8 -84.5597839355469
13.2 -84.5597839355469
nan nan
13 -84.5597839355469
13 -83.8401641845703
nan nan
12.8 -83.8401641845703
13.2 -83.8401641845703
};
\addplot [line width=0.9pt, gray]
table {%
13.8 -51.2292442321777
14.2 -51.2292442321777
nan nan
14 -51.2292442321777
14 -50.8226737976074
nan nan
13.8 -50.8226737976074
14.2 -50.8226737976074
};
\path [draw=contrabarcolor, thick, dash pattern=on 1.5pt off 2.475pt]
(axis cs:4,-42.3)
--(axis cs:14.5,-42.3);

\path [draw=borelcolor, thick, dash pattern=on 1.5pt off 2.475pt]
(axis cs:1,-35.4)
--(axis cs:14.5,-35.4);

\draw (axis cs:1.5,-47) node[
  scale=0.8,
  anchor=west,
  text=contrabarcolor,
  rotate=0.0
]{ ContraBAR};
\draw (axis cs:-0.5,-40) node[
  scale=0.8,
  anchor=west,
  text=borelcolor,
  rotate=0.0
]{ BOReL};
\end{axis}

\end{tikzpicture}
    \end{minipage}
    \caption{Left: Performance on 2D-Goal when using a noisy ground-truth HiP, with a uniformly random HiP being chosen with probability $\sigma$ to achieve a desired $P(\tilde{z}=z)$. Right: Performance of our method on 1D-Goal with varying randomness of the HiP transition. Both show means and 95\% CI across 10 trials.}
    \label{fig:xynoisyxlrnoisy}
\end{figure}
It is interesting that BOReL performs relatively poor in the 2D-Goal task, while achieving a relatively high linear probe accuracy.
To investigate this finding, we perfomed an experiment in which we simulate a noisy Oracle that returns the correct HiP with probability $P(\tilde{z}=z)$ and else returns a uniformly random different HiP. The results in Fig. \ref{fig:xynoisyxlrnoisy} (left) show that even with a relatively high accuracy $P(\tilde{z}=z)=0.84$ the noise can lead to significantly worse results. This is consistent with results in \cite{yang_rorl_2022,yang_towards_2024}, showing the sensitivity of Offline RL to state noise.

We are also interested in the question of when our method should be prefered over Bayes Adaptive approaches such as BOReL and ContraBAR.
Intuitively, our method should perform better when the next HiP can be accurately predicted, while Bayes Adaptive methods should be preferable if it is less predictable, i.e. closer to a HiP-MDP than a DP-MDP. In Fig. \ref{fig:xynoisyxlrnoisy} (right) we show the attained reward when the HiP transition function $P(z_i|z_{1:i-1})$ is changed to a uniformly random HiP with probability $\sigma$.
The results align with our intuition of our method performing better than BOReL when the HiP is predictable, but worse when it becomes less predictable.

\section{Conclusion}
We investigated a novel problem setting in Offline Reinforcement Learning, in which the training data is generated from multiple deployments with non-stationary transition and reward functions.
The problem is formulated as multiple rollouts of a Dynamic-Parameter MDP, a Hidden-Parameter MDP (HiP-MDP) in which the HiP evolves across trajectories.
We proposed a method using contrastive learning that learns a representation of the HiP, predicts the HiP during evaluation and trains a policy conditioned on it.
We showed that our method is able to learn a useful representation of the HiP, allowing us to train a policy that often performs better than baseline methods in experiments.

\section*{Acknowledgements}
J.A. was supported by the Microsoft Research Asia D-CORE program. T.O. was partially supported by JSPS KAKENHI Grant Number JP23K18476.

\appendix
\section{Further Related Work}
We will here discuss additional related work, first on Nonstationary RL, then Multi-Task Offline RL and finally Contrastive Learning in RL.
\paragraph{Nonstationary RL}
The perhaps closest related work has been presented by \citet{xie_deep_2021}, in which they propose the DP-MDP formulation which we use in our problem setting. Unlike us, their work focuses on an online lifelong-RL setting with a single deployment for both training and evaluation. They also proposed the method LILAC, an offline variation of which we use as a baseline.
\cite{chen_adaptive_2022} address non-stationary environments with piece-wise stable context. They address a setting where parts of each episode are stationary, but do not consider structure between tasks.
\citet{wang_robust_2023} derive a robust method to test whether nonstationarity occurs in a given offline RL dataset. Their work focuses on identifying whether such non-stationarity occurs, while we focus on how we can learn a policy that adapts to it.
\cite{yin_towards_2021} provide a theoretical investigation of nonstationary Offline RL in a setting where the nonstationary occurs during each episode but not between episodes, while in our setting it occurs between episodes and not within each episode.
\citet{dulac-arnold_challenges_2019} also discusses non-stationarity introduced by wear and tear and proposes environments to investigate these challenges in online RL.
\citet{chandak_off-policy_2022} consider off-policy evaluation in a generalization of the DP-MDP in which the next HiP depends on the previous HiPs and the actions taken within the previous episode. 
Off-policy evaluation is an important step towards Offline RL, but it is not immediately clear how to extend their method to the control setting or to complex environments which we address in our work.

\paragraph{Multi-Task Offline RL}
A related setting to ours is Multi-Task Offline RL.
The main difference is that in Multi-Task RL we are given the task identity during training and therefore do not have to learn an inference model in an unsupervised setting.
\citet{li_multi-task_2020} and \citet{li_focal_2022}  propose methods to address such a setting where the task ID is given during training and can therefore not be applied to our setting.
\citet{liu_dara_2022} and \citet{xue_state_2023} address the Offline Transfer Learning setting, in which a large amount of source domain data is available but limited target domain data with different dynamics.

\paragraph{Contrastive Learning}
CPC \citep{oord_representation_2019} has been used in RL before, the original authors themselves used it as an auxiliary loss in Atari tasks \citep{oord_representation_2019}.
ContraBAR \citep{choshen_contrabar_2023} proposed to use CPC to learn Bayes-Optimal policies in the Bayes adaptive RL setting \citep{Zintgraf2020}. 
Aside from CPC, other contrastive learning approaches have been applied in RL, for representation learning \citep{kipf_contrastive_2019,srinivas_curl_2020,van_der_pol_plannable_2020},  or for domain inference networks in multi-task RL \citep{li_focal_2022,lan_contrastive_2023}.

\bibliography{mybib}
\bibliographystyle{rlc}

\newpage
\section{Theoretical Analysis of CPC Inference}
\label{app:analysisofcpc}
For completeness, we show our derivation which follows the original CPC paper closely \citep{oord_representation_2019} and is related to those by \cite{choshen_contrabar_2023}, which studies CPC in a Bayes-Adaptive MDP setting.

One difference to note between our setting and that considered in \cite{choshen_contrabar_2023}, is the change of the hidden-parameter between episodes and therefore between inputs to the CPC model.
While in their setting the hidden-parameter is the same for all inputs (transitions), in our setting it evolves between inputs (trajectories).
We provide a full illustration of the data generation graphical model and CPC model in Fig. \ref{fig:appdatamodel}:
\begin{figure}[h]
    \centering
    \includegraphics[width=0.8\linewidth]{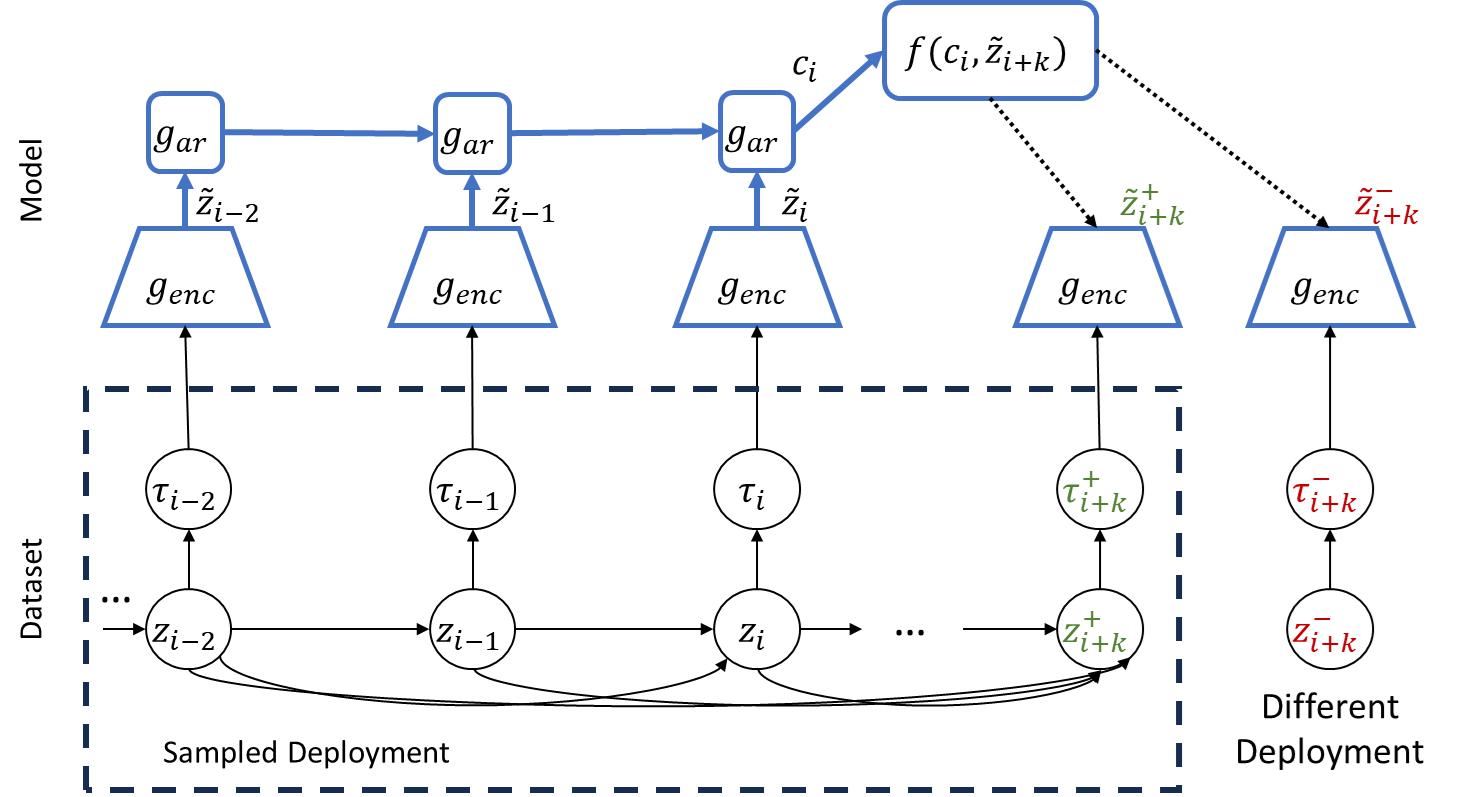}
    \caption{Illustration of the data generation and our model.}
    \label{fig:appdatamodel}
\end{figure}

To show that CPC maximizes the mutual information between $c_i$ and $\tau_{i+k}$, we use the same derivation as 2.3 and A.1 of \cite{oord_representation_2019}:
\begin{lemma}
Let the InfoNCE loss in equation \ref{eq:cpcreprloss} be jointly minimized by $f, g_{\mathrm{enc}}, g_{\mathrm{ar}}$, then for any trajectory $\tau$, with $c_i=g_{\mathrm{ar}}(g_{\mathrm{enc}}(\tau_{i-1},\tau_{i-2},\dots, \tau_{1}))$, we have
$$f(\tau_{i+k},c_i)\propto \frac{P(\tau_{i+k}|c_i)}{P(\tau_{i+k})}$$ 
\end{lemma}
\begin{proof}
This follows from the InfoNCE loss \eqref{eq:cpcreprloss} being the categorical cross-entropy of correctly classifying the positive sample in the given batch of samples $B = (\tau_{i+k}^{1},\dots,\tau_{i+k}^{N})$.
Following \cite{oord_representation_2019}, we can write the classification problem as learning a classifier $P(d=o|B,c_i)$ with $[d=o]$ being the indicator of the positive sample (omitting the subscript $i+k$ for all $\tau$):
\begin{equation}
\begin{aligned}
    p(d=o|B,c_i) &= \frac
    {p(\tau^o|c_i) \prod_{l\not=o}p(\tau^l)}
    {\sum_{j=1}^{N} p(\tau^j|c_i) \prod_{l\not=j}p(\tau^l)} \\
    &= \frac
    {
        \frac{p(\tau^o|c_i)}{p(\tau^o)}
    }
    {
        \sum_{j=1}^{N} \frac{p(\tau^j|c_i)}{p(\tau^j)} 
    }\,,
\end{aligned}
\end{equation}
which we can see is proportional to $\frac{P(\tau_{i+k}|c_i)}{P(\tau_{i+k})}$.
As such, the optimal value of the classification problem learned by $f(\tau_{i+k},c_i)$ is proportional to it as well.
\end{proof}

\begin{lemma}
    Let the InfoNCE loss in equation \ref{eq:cpcreprloss} be jointly minimized by $f, g_{\mathrm{enc}}, g_{\mathrm{ar}}$, then for any trajectory $\tau$, with $c=g_{\mathrm{ar}}(g_{\mathrm{enc}}(\tau_{i-1},\tau_{i-2},\dots,\tau_1)$, we have
    $$\mathcal{L}_{repr} \geq \log(N-1) - I(\tau_{i+k};c_i) $$
\end{lemma}
\begin{proof}
    Let $B_\mathrm{neg}$ be the negative samples in a batch $B$. By inserting the optimal value of $f(\tau_{i+k},c_i)$ into the loss function \eqref{eq:cpcreprloss}, we get (omitting the subscript $i+k$ for all $\tau$)
    \begin{equation}
        \begin{aligned}
            \mathcal{L}_{repr} &=-\mathbb{E} \log\left[ \frac
            {\frac{p(\tau^+|c_i)}{p(\tau^+)}}
            {\frac{p(\tau^+|c_i)}{p(\tau^+)} + \sum_{\tau^{-} \in B_\mathrm{neg}}\frac{p(\tau^-|c_i)}{p(\tau^-)}}
            \right]  \\
            &=\mathbb{E}\log\left[ 
            {1 + \frac{p(\tau^+)}{p(\tau^+|c_i)} \sum_{\tau^{-} \in B_\mathrm{neg}}\frac{p(\tau^-|c_i)}{p(\tau^-)}}
            \right] \\
            &\approx \mathbb{E} \log\left[
            {1 + \frac{p(\tau^+)}{p(\tau^+|c_i)} (N-1) \mathbb{E}_{\tau^-}\frac{p(\tau^-|c_i)}{p(\tau^-)}}
            \right] \\
            &=\mathbb{E} \log\left[
            {1 + \frac{p(\tau^+)}{p(\tau^+|c_t)} (N-1)}
            \right] \\
            &\geq \mathbb{E} \log\left[
            {\frac{p(\tau^+)}{p(\tau^+|c_t)} (N-1)}
            \right] \\
            &= H(\tau^+ | c_i) - H(\tau^+) + \log(N-1) \\
            &= - I(\tau^+;c_i) + \log(N-1) \,.
        \end{aligned}
    \end{equation} 
\end{proof}

\section{Implementation Details}
\label{app:implementation_details}
Our implementation is available on github, additional implementation details can be found there.
Note that we smooth the reward plots with a moving window of size 3 for clarity.
Confidence intervals and means are calculated across trials consisting of representation learning and offline RL with a fixed dataset for all trials.

\subsection{Environments}
We use the same DP-MDPs and thus the same sets of HiPs in training and evaluation.

\subsubsection{Low-Dimensional Tasks}
In 1D-Goal, the state space is $S=[-2,2]$ and action space is $A=[-0.1,0.1]$.
The initial state is uniformly sampled from $S$ and the goal g is at $g=z\in\{-1,1\}$, depending on the hidden parameter $z$.
The deterministic transition function is $s_{t+1} = s_t + a_t$ and the reward is the negative absolute distance to the goal location $r_t= -|s_t - g|$.
The hidden parameter deterministically switches each episode, i.e. $P(z_{i+1}=-1|z_{i}=1)=P(z_{i+1}=1|z_{i}=-1)=1$, with each deployment consisting of 10 episodes.
For the experiment in Fig. \ref{fig:xynoisyxlrnoisy} (right), we change the transition function to $P(z_{i+1}=1|z_{i}=-1)=P(z_{i+1}=-1|z_{i}=1)=1-\sigma / 2$.

Our 2D-tasks are a continuous environment with action space $S=[-2,2]^2$ and action space $A=[-0.1,0.1]^2$.
The transition function is deterministic and follows $s_{t+1} = s_t + a_t$.
The goal is to navigate to a goal location.
In \textit{2D Goal}, the agent starts at a uniformly random location and the reward function is the negative Euclidean distance to the goal location $r_t= -|s_t - g|_2$.
The goal location $g$ lies on the unit circle and is determined by the hidden-parameter $z$: $g(z)=(\sin(z), \cos(z))^\intercal$.
The hidden-parameter follows a triangle-wave with period 8 with $z \in [0, \frac{3}{2}\pi]$, i.e.$z\in\{0,\frac{3}{8}\pi,\frac{6}{8}\pi,\frac{9}{8}\pi,\frac{3}{2}\pi\}$ , and each deployment is 20 episodes long.
This task is similar to "Semi-Circle" in \cite{dorfman_offline_2021_borel}.

In \textit{2D Wind}, the agent always starts at the origin $(0,0)^\intercal$ and the goal location is always $g=(1,0)^\intercal$.
Here we use a sparse unit reward $r_t=1$ if the distance $|s_t - g|_2 < 0.2$, else no reward is given $r_t=0$.
The dynamics are changed to include a disturbance, $s_{t+1}=s_t + a_t + 0.09 (\sin(z), \cos(z))^\intercal$, where $z$ follows a sawtooth-wave with period five on $[0,2\pi]$, i.e. $z\in\{(0,\frac{2}{5}\pi,\frac{4}{5}\pi,\frac{6}{5}\pi, \frac{8}{5}\pi\}$, and each deployment is 20 episodes long.
This task is based on "Wind" in \cite{dorfman_offline_2021_borel}.

\subsubsection{Ant}
We modify the Ant environment provided in Brax \citep{freeman_brax_2021} to allow for multiple different robot configurations.
We keep the original reward function, which rewards the robot for forward movement.

In \textit{Ant-Weight}, we modify the mass of the base (the sphere) of the robot by multiplying it by the current hidden-parameter $z$.
For the HiP evolution we here use a sawtooth-wave with period 5 and $z \in [0.5,2.5]$, i.e. $z\in\{0.5, 1.0, 1.5, 2.0, 2.5\}$, with each deployment lasting 20 episodes.
We use the default observation.

In \textit{Ant-Leg} we multiply the length of each leg, both the "femur" and "tibia", by the hidden-parameter $z$.
We again use a sawtooth-wave with period 5 and $z \in [0.75, 1.25]$, i.e. $z\in\{0.75, 0.875, 1.0, 1.125, 1.25\}$, with 20 episodes per deployment.
We alter the observation by removing the z-component of the position of the base, as it otherwise directly represents the hidden-parameter in the first timestep.

\subsubsection{Barkour}
Unlike for the Ant experiment, for Barkour we now use Mujoco-MJX to simulate the robot and modify the environment provided in the Mujoco-MJX Tutorial.\footnote{\url{https://github.com/google-deepmind/mujoco/blob/721e2d5589d3fdafd440009374a31521214088b7/mjx/tutorial.ipynb}}
The Barkour environment usually trains the robot to track an input command linear velocity and angular velocity, which is sampled uniformly during training.
As this sampling adds a high amount of variance to reward function, we change the reward-function by setting a constant desired forward velocity and removing the reward for angular velocity tracking.
We use this altered reward to finetune a pretrained Barkour policy and then also to collect the dataset and evaluate the Offline RL policy.
To simulate the robot having to carry a varying load, we alter the weight of the chassis by multiplying it with the hidden-parameter, following a sawtooth-wave with period 5 in $z\in[1.0,4.0]$, i.e. $z\in\{1.0, 1.75, 2.5, 3.25, 4.0\}$, for 15 episodes per deployment.
This task is inspired by "minitaur-payload" in \cite{xie_deep_2021}.

\section{Method Details}
We outline the main implementation details here, while additional details can be found in the implementation on github.
We use Adam \citep{Kingma2014} in all experiments.
\subsection{CPC Implementation}
We implement $g_\mathrm{enc}$ as a two layer MLP with 128 units per layer and ReLU activations and $g_\mathrm{ar}$ as a GRU with 16 units in the low-dimensional and 32 units in the high-dimensional tasks.
The classifier $f$ is implemented as an MLP with two hidden layers with 128 units per layer and ReLU activations. 
As the initial parameters performed well across tasks, we did not perform extensive grid-search as for the baselines, and only searched over latent dimensionality $\{2,4,6,8\}$ per task.

The positive and negative samples are sampled as follows:
To obtain positive and negative trajectories we sample one positive deployment $d^+=(\tau_1^+,\dots,\tau_M^+)$, $N^-$ negative deployments $\{d^{-,j}\}_{j=1}^{N^-}$, with $d^{-,j}=(\tau_1^{-,j},\dots,\tau_M^{-,j})$ and deployment step $i\sim U([N_\mathrm{CPC},M-k])$.
Then we obtain the context encoding $c_i=g_\mathrm{ar}(g_\mathrm{enc}(\tau_{i}),\dots,g_\mathrm{enc}(\tau_{i-N_\mathrm{CPC}}))$ and use it as input to $f$ with positive sample $\tau_{i+k}^+$ and negative samples $\{\tau_{i+k}^{-,j}\}_{j=1}^N$.

\subsection{Bayesian RL Baseline Implementation}
We implement our Bayesian RL baselines based on VariBAD \citep{Zintgraf2020}/BOReL \citep{dorfman_offline_2021_borel} and ContraBAR \citep{choshen_contrabar_2023}.
We note that while they use two episodes to measure the performance of the agent, we measure the reward over a single episode due to hidden-parameter changing after each episode in our setting.
We truncate the trajectory length used for HiP inference to 10 in Ant-Leg and Barkour-Weight and 50 in Ant-Weight, to allow for efficient training.
We freeze the last inferred latent and use it for the remainder of the episode.

\subsubsection{BOReL Implementation}
We use a recurrent encoder consisting of two fully connected layers with RELU activations, followed by a GRU, followed by a fully connected layer each to output $\mu$ and $\Sigma$.
The reward and transitions decoders each consist of two or three hidden layers with ReLU activations, where the latent is input into the second layer.
We only use either the reward or transition decoder depending on whether the reward or transition function changes in an environment.
We found RL training to perform better when only conditioning on the mean $\mu$ without the covariance matrix $\Sigma$.
We perform a grid-search over representation hyperparameters as shown in the Table \ref{tab:borelvrnngrid} on 1D-Goal, 2D-Goal, 2D-Wind, Ant-Weight and Ant-Leg.
Due to resource constraints we do not perform a full grid-search on barkour, but did our best effort to choose well-performing hyperaprameters and evaluated different choices for the latent-dimensionality.

\begin{table}[]
    \centering
    \begin{tabular}{c|c}
        Hyperparameter & Gridsearch Values BORel, VRNN \\
        \hline
        Hidden Units Decoder/Encoder &  $\{128, 256\}$ \\
        Decoder Hidden Layers & $\{2,3\}$ \\
        VAE $\beta$ & $\{10^{-4}, 10^{-5}, 10^{-6}, 10^{-7}, 0\}$ \\
        Latent Dimensionality & $\{4,8\}$
    \end{tabular}
    \caption{Hyperparameters values used in Grid-Search for BOReL and VRNN}
    \label{tab:borelvrnngrid}
\end{table}

\subsubsection{ContraBAR Implementation}
For ContraBAR we found the network structure of encoder and classifier to strongly impact the achieved performance.
For fairness of comparison we evaluated two different structures: Firstly, a network structure as proposed by \cite{choshen_contrabar_2023} with separate encoders for reward, state, and action, a GRU with larger dimensionality $d=64$ and a small, single-layer classifier $f$. Secondly, we evaluated the same structure as used in our approach, with more powerful fully-connected encoders and classifiers, but smaller GRU dimensionality $d=4,8$.
We found the latter structure to perform better on the 2D tasks, while the former usually resulted in better representations in the other tasks.
As recommended by \cite{choshen_contrabar_2023}, we use the Action-GRU only in tasks where the transition function changes, omitting it otherwise.
We performed a gridsearch to optimize the hyperparameters over the values shown in Table \ref{tab:contrabargrid} on 1D-Goal, 2D-Goal, 2D-Wind, Ant-Weight and Ant-Leg.
Due to resource constraints we do not perform a full grid-search on Barkour, but did our best effort to choose well-performing hyperaprameters and evaluated different choices for the latent-dimensionality.

\begin{table}[]
    \centering
    \begin{tabular}{c|c}
        Hyperparameter & Gridsearch Values ContraBAR \\
        \hline
        Hidden Units Decoder/Encoder &  $\{64, 128, 256\}$ \\
        Encoder Architecture & $\{$SplitEnc, MLPEnc$\}$ \\
        Encoder Hidden Layers & $\{1,2\}$ \\
        Latent Dimensionality & $\{2,4,8\}$
    \end{tabular}
    \caption{Hyperparameters values used in Grid-Search for ContraBAR}
    \label{tab:contrabargrid}
\end{table}

\subsection{VRNN Implementation}
We evaluated different approaches to creating an offline variant of LILAC with different network architectures and training mechanisms, and report the most successful one we found.
As it is different in network structure and training method we refer to it simply as \textit{VRNN} in our experiments.
We base our implementation on a VRNN \citep{chung_recurrent_2016VRNN}, a type of dynamic VAE with an LSTM prior.
The encoder is a two layer MLP with ReLU activations, the observation and latent feature extractors are implemented as fully connected layers with ReLU activations and the learned prior is implemented an LSTM followed by a fully connected layer with ReLU activations and two linear output layers for $\mu, \Sigma$.
The decoders are MLPs with two or three layers and ReLU activations.
We train the VRNN by sampling two different transitions from each episode, where one transition is used in the encoder and a different transition is used in the decoder, to prevent the encoder from simply learning the state or reward functions.
We had issues with innacurate predictions and thus, during evaluation, we condition the policy on a sampled latent $\tilde{z}$ of an episode from our dataset with the same HiP $z$.
This method can therefore be considered a semi-oracle that uses oracle information during evaluation but not during training.
As in BOReL, we found RL training to perform better when only conditioning on the mean $\mu$ without the covariance matrix $\Sigma$.
Hyperparameters are optimized by gridsearch over values shown in Table \ref{tab:borelvrnngrid} on 1D-Goal, 2D-Goal, 2D-Wind, Ant-Weight and Ant-Leg.
Due to resource constraints we do not perform a full grid-search on Barkour, but did our best effort to choose well-performing hyperaprameters and evaluated different choices for the latent-dimensionality.

\subsection{Offline Reinforcement Learning Implementation}
On each task, we use the same network structure and RL hyper-parameters for all methods. We do deviate from the original parameters of TD3+BC \citep{fujimoto_minimalist_2021} as shown in Table \ref{tab:TD3BCHyperparam}.

\begin{table}[h]
    \centering
    \begin{tabular}{cccccc}
         Hyperparameter & 1D-Goal & 2D-Win & 2D-Goal & Ant-Weight, Ant-Leg & Barkour \\
         \hline
         Critic Network & [256, 256] & [256, 256] & [256, 256] & [128, 128] & [128, 128, 128] \\
         Policy Network & [256, 256] & [256, 256] & [256, 256] & [128, 128] & [128, 128, 128] \\
         Layer Norm & Yes & Yes & No & Yes & Yes \\
         BC $\lambda$ & 2.5 & 6.5 & 6.5 & 6.5 & 6.5 \\
         Learning Rate & $3\cdot 10^{-4}$ & $3\cdot 10^{-4}$ & $1\cdot 10^{-3}$ & $1\cdot 10^{-3}$ & $1\cdot 10^{-3}$ \\
         Batch Size & 512 & 512 & 512 & 512 & 512 \\
    \end{tabular}
    \caption{TD3 BC Hyperparameters}
    \label{tab:TD3BCHyperparam}
\end{table}

\section{RLiable Visualization}
To provide more informative evaluation metrics than the mean reward values reported in the main text, we provide additional visualizations as suggested by \cite{agarwal_deep_2022}, using the RLiable package.

\subsection{1D-Goal}
\begin{tikzpicture}

\definecolor{chocolate213940}{RGB}{213,94,0}
\definecolor{darkcyan1115178}{RGB}{1,115,178}
\definecolor{darkcyan2158115}{RGB}{2,158,115}
\definecolor{darkgray176}{RGB}{176,176,176}
\definecolor{darkorange2221435}{RGB}{222,143,5}
\definecolor{orchid204120188}{RGB}{204,120,188}
\definecolor{peru20214597}{RGB}{202,145,97}

\begin{groupplot}[
group style={
group size=3 by 1,
y descriptions at=edge left,
}, 
width=\linewidth/3.3, 
]\nextgroupplot[
tick align=outside,
tick pos=left,
title={Median},
x grid style={darkgray176},
xmajorgrids,
xmin=-52, xmax=-15.1606496654749,
xtick style={color=black},
y grid style={darkgray176},
ymin=-0.58, ymax=5.58,
ytick style={color=black},
ytick={0,1,2,3,4,5},
yticklabels={ContraBAR,BOReL,VRNN,COSPA,Blind,Oracle}
]
\draw[draw=none,fill=darkcyan1115178,fill opacity=0.75] (axis cs:-46.4358435993195,-0.3) rectangle (axis cs:-38.5802365059852,0.3);
\addlegendimage{ybar,ybar legend,draw=none,fill=darkcyan1115178,fill opacity=0.75}

\draw[draw=none,fill=darkorange2221435,fill opacity=0.75] (axis cs:-41.1120004091263,0.7) rectangle (axis cs:-32.0258815088272,1.3);
\addlegendimage{ybar,ybar legend,draw=none,fill=darkorange2221435,fill opacity=0.75}

\draw[draw=none,fill=darkcyan2158115,fill opacity=0.75] (axis cs:-25.9656824464798,1.7) rectangle (axis cs:-18.4755211640596,2.3);
\addlegendimage{ybar,ybar legend,draw=none,fill=darkcyan2158115,fill opacity=0.75}

\draw[draw=none,fill=chocolate213940,fill opacity=0.75] (axis cs:-21.6241189143658,2.7) rectangle (axis cs:-16.8628368269205,3.3);
\addlegendimage{ybar,ybar legend,draw=none,fill=chocolate213940,fill opacity=0.75}

\draw[draw=none,fill=orchid204120188,fill opacity=0.75] (axis cs:-50.9850216665268,3.7) rectangle (axis cs:-50.951985490799,4.3);
\addlegendimage{ybar,ybar legend,draw=none,fill=orchid204120188,fill opacity=0.75}

\draw[draw=none,fill=peru20214597,fill opacity=0.75] (axis cs:-23.3558299561739,4.7) rectangle (axis cs:-18.6901208299398,5.3);
\addlegendimage{ybar,ybar legend,draw=none,fill=peru20214597,fill opacity=0.75}

\path [draw=black, draw opacity=0.5, semithick]
(axis cs:-42.4918267822266,-0.28125)
--(axis cs:-42.4918267822266,0.225);

\path [draw=black, draw opacity=0.5, semithick]
(axis cs:-36.4831686019897,0.71875)
--(axis cs:-36.4831686019897,1.225);

\path [draw=black, draw opacity=0.5, semithick]
(axis cs:-21.8917444372177,1.71875)
--(axis cs:-21.8917444372177,2.225);

\path [draw=black, draw opacity=0.5, semithick]
(axis cs:-18.6516311597824,2.71875)
--(axis cs:-18.6516311597824,3.225);

\path [draw=black, draw opacity=0.5, semithick]
(axis cs:-50.9684561347961,3.71875)
--(axis cs:-50.9684561347961,4.225);

\path [draw=black, draw opacity=0.5, semithick]
(axis cs:-20.8858898639679,4.71875)
--(axis cs:-20.8858898639679,5.225);

\nextgroupplot[
tick align=outside,
tick pos=left,
title={IQM},
x grid style={darkgray176},
xmajorgrids,
xmin=-52, xmax=-15.1606496654749,
xtick style={color=black},
ymin=-0.58, ymax=5.58
]
\draw[draw=none,fill=darkcyan1115178,fill opacity=0.75] (axis cs:-46.8250473537445,-0.3) rectangle (axis cs:-37.7202789239883,0.3);
\draw[draw=none,fill=darkorange2221435,fill opacity=0.75] (axis cs:-41.4311135864258,0.7) rectangle (axis cs:-31.1866861038208,1.3);
\draw[draw=none,fill=darkcyan2158115,fill opacity=0.75] (axis cs:-19.2065657310486,1.7) rectangle (axis cs:-16.2158682365417,2.3);
\draw[draw=none,fill=chocolate213940,fill opacity=0.75] (axis cs:-17.9395558338165,2.7) rectangle (axis cs:-16.5985462806225,3.3);
\draw[draw=none,fill=orchid204120188,fill opacity=0.75] (axis cs:-50.9885733423233,3.7) rectangle (axis cs:-50.9504091939926,4.3);
\draw[draw=none,fill=peru20214597,fill opacity=0.75] (axis cs:-20.9576041603088,4.7) rectangle (axis cs:-16.9959100549221,5.3);
\path [draw=black, draw opacity=0.5, semithick]
(axis cs:-42.2603467559814,-0.28125)
--(axis cs:-42.2603467559814,0.225);

\path [draw=black, draw opacity=0.5, semithick]
(axis cs:-36.5046291732788,0.71875)
--(axis cs:-36.5046291732788,1.225);

\path [draw=black, draw opacity=0.5, semithick]
(axis cs:-16.8149101161957,1.71875)
--(axis cs:-16.8149101161957,2.225);

\path [draw=black, draw opacity=0.5, semithick]
(axis cs:-17.2838923549652,2.71875)
--(axis cs:-17.2838923549652,3.225);

\path [draw=black, draw opacity=0.5, semithick]
(axis cs:-50.9705621337891,3.71875)
--(axis cs:-50.9705621337891,4.225);

\path [draw=black, draw opacity=0.5, semithick]
(axis cs:-18.6373333263397,4.71875)
--(axis cs:-18.6373333263397,5.225);

\nextgroupplot[
tick align=outside,
tick pos=left,
title={Mean},
x grid style={darkgray176},
xmajorgrids,
xmin=-52, xmax=-15.1606496654749,
xtick style={color=black},
ymin=-0.58, ymax=5.58
]
\draw[draw=none,fill=darkcyan1115178,fill opacity=0.75] (axis cs:-46.4358435993195,-0.3) rectangle (axis cs:-38.5802365059852,0.3);
\draw[draw=none,fill=darkorange2221435,fill opacity=0.75] (axis cs:-41.1120004091263,0.7) rectangle (axis cs:-32.0258815088272,1.3);
\draw[draw=none,fill=darkcyan2158115,fill opacity=0.75] (axis cs:-25.9656824464798,1.7) rectangle (axis cs:-18.4755211640596,2.3);
\draw[draw=none,fill=chocolate213940,fill opacity=0.75] (axis cs:-21.6241189143658,2.7) rectangle (axis cs:-16.8628368269205,3.3);
\draw[draw=none,fill=orchid204120188,fill opacity=0.75] (axis cs:-50.9850216665268,3.7) rectangle (axis cs:-50.951985490799,4.3);
\draw[draw=none,fill=peru20214597,fill opacity=0.75] (axis cs:-23.3558299561739,4.7) rectangle (axis cs:-18.6901208299398,5.3);
\path [draw=black, draw opacity=0.5, semithick]
(axis cs:-42.4918267822266,-0.28125)
--(axis cs:-42.4918267822266,0.225);

\path [draw=black, draw opacity=0.5, semithick]
(axis cs:-36.4831686019897,0.71875)
--(axis cs:-36.4831686019897,1.225);

\path [draw=black, draw opacity=0.5, semithick]
(axis cs:-21.8917444372177,1.71875)
--(axis cs:-21.8917444372177,2.225);

\path [draw=black, draw opacity=0.5, semithick]
(axis cs:-18.6516311597824,2.71875)
--(axis cs:-18.6516311597824,3.225);

\path [draw=black, draw opacity=0.5, semithick]
(axis cs:-50.9684561347961,3.71875)
--(axis cs:-50.9684561347961,4.225);

\path [draw=black, draw opacity=0.5, semithick]
(axis cs:-20.8858898639679,4.71875)
--(axis cs:-20.8858898639679,5.225);

\end{groupplot}

\end{tikzpicture}

\input{pgf/Rliable/xlrevolvev3perfprof}

\subsection{2D-Goal}
\begin{tikzpicture}

\definecolor{chocolate213940}{RGB}{213,94,0}
\definecolor{darkcyan1115178}{RGB}{1,115,178}
\definecolor{darkcyan2158115}{RGB}{2,158,115}
\definecolor{darkgray176}{RGB}{176,176,176}
\definecolor{darkorange2221435}{RGB}{222,143,5}
\definecolor{orchid204120188}{RGB}{204,120,188}
\definecolor{peru20214597}{RGB}{202,145,97}

\begin{groupplot}[
group style={
group size=3 by 1,
y descriptions at=edge left,
}, 
width=\linewidth/3.3, 
]\nextgroupplot[
tick align=outside,
tick pos=left,
title={Median},
x grid style={darkgray176},
xmajorgrids,
xmin=-85.2040078420639, xmax=-27.1434605512857,
xtick style={color=black},
y grid style={darkgray176},
ymin=-0.58, ymax=5.58,
ytick style={color=black},
ytick={0,1,2,3,4,5},
yticklabels={ContraBAR,VRNN,BOReL,COSPA,Oracle,Blind}
]
\draw[draw=none,fill=darkcyan1115178,fill opacity=0.75] (axis cs:-76.0900220327377,-0.3) rectangle (axis cs:-62.7650328969955,0.3);
\addlegendimage{ybar,ybar legend,draw=none,fill=darkcyan1115178,fill opacity=0.75}

\draw[draw=none,fill=darkorange2221435,fill opacity=0.75] (axis cs:-60.4493686628342,0.7) rectangle (axis cs:-52.6118664102554,1.3);
\addlegendimage{ybar,ybar legend,draw=none,fill=darkorange2221435,fill opacity=0.75}

\draw[draw=none,fill=darkcyan2158115,fill opacity=0.75] (axis cs:-53.6115662593842,1.7) rectangle (axis cs:-50.8277456378937,2.3);
\addlegendimage{ybar,ybar legend,draw=none,fill=darkcyan2158115,fill opacity=0.75}

\draw[draw=none,fill=chocolate213940,fill opacity=0.75] (axis cs:-39.673004755497,2.7) rectangle (axis cs:-33.8933067359924,3.3);
\addlegendimage{ybar,ybar legend,draw=none,fill=chocolate213940,fill opacity=0.75}

\draw[draw=none,fill=orchid204120188,fill opacity=0.75] (axis cs:-33.159646894455,3.7) rectangle (axis cs:-29.9133306398392,4.3);
\addlegendimage{ybar,ybar legend,draw=none,fill=orchid204120188,fill opacity=0.75}

\draw[draw=none,fill=peru20214597,fill opacity=0.75] (axis cs:-52.5193189706802,4.7) rectangle (axis cs:-52.2459249725342,5.3);
\addlegendimage{ybar,ybar legend,draw=none,fill=peru20214597,fill opacity=0.75}

\path [draw=black, draw opacity=0.5, semithick]
(axis cs:-68.9334267044067,-0.28125)
--(axis cs:-68.9334267044067,0.225);

\path [draw=black, draw opacity=0.5, semithick]
(axis cs:-56.1219903945923,0.71875)
--(axis cs:-56.1219903945923,1.225);

\path [draw=black, draw opacity=0.5, semithick]
(axis cs:-52.0833439254761,1.71875)
--(axis cs:-52.0833439254761,2.225);

\path [draw=black, draw opacity=0.5, semithick]
(axis cs:-36.7584313201904,2.71875)
--(axis cs:-36.7584313201904,3.225);

\path [draw=black, draw opacity=0.5, semithick]
(axis cs:-31.4792595672607,3.71875)
--(axis cs:-31.4792595672607,4.225);

\path [draw=black, draw opacity=0.5, semithick]
(axis cs:-52.3597022247314,4.71875)
--(axis cs:-52.3597022247314,5.225);

\nextgroupplot[
tick align=outside,
tick pos=left,
title={IQM},
x grid style={darkgray176},
xmajorgrids,
xmin=-71.9869474582672, xmax=-27.577901281786,
xtick style={color=black},
ymin=-0.58, ymax=5.58
]
\draw[draw=none,fill=darkcyan1115178,fill opacity=0.75] (axis cs:-71.9869474582672,-0.3) rectangle (axis cs:-60.4357608852386,0.3);
\draw[draw=none,fill=darkorange2221435,fill opacity=0.75] (axis cs:-57.6243222446442,0.7) rectangle (axis cs:-51.9612874450684,1.3);
\draw[draw=none,fill=darkcyan2158115,fill opacity=0.75] (axis cs:-52.8251791419983,1.7) rectangle (axis cs:-50.4165114994049,2.3);
\draw[draw=none,fill=chocolate213940,fill opacity=0.75] (axis cs:-40.2829750003815,2.7) rectangle (axis cs:-33.0708066892624,3.3);
\draw[draw=none,fill=orchid204120188,fill opacity=0.75] (axis cs:-32.9834794483185,3.7) rectangle (axis cs:-29.6926177663803,4.3);
\draw[draw=none,fill=peru20214597,fill opacity=0.75] (axis cs:-52.3747457504273,4.7) rectangle (axis cs:-52.226346906662,5.3);
\path [draw=black, draw opacity=0.5, semithick]
(axis cs:-64.7830792999268,-0.28125)
--(axis cs:-64.7830792999268,0.225);

\path [draw=black, draw opacity=0.5, semithick]
(axis cs:-55.1974265289307,0.71875)
--(axis cs:-55.1974265289307,1.225);

\path [draw=black, draw opacity=0.5, semithick]
(axis cs:-51.6018504333496,1.71875)
--(axis cs:-51.6018504333496,2.225);

\path [draw=black, draw opacity=0.5, semithick]
(axis cs:-36.8808036804199,2.71875)
--(axis cs:-36.8808036804199,3.225);

\path [draw=black, draw opacity=0.5, semithick]
(axis cs:-31.254627456665,3.71875)
--(axis cs:-31.254627456665,4.225);

\path [draw=black, draw opacity=0.5, semithick]
(axis cs:-52.2937219238281,4.71875)
--(axis cs:-52.2937219238281,5.225);

\nextgroupplot[
tick align=outside,
tick pos=left,
title={Mean},
x grid style={darkgray176},
xmajorgrids,
xmin=-76.0900220327377, xmax=-27.6044960701943,
xtick style={color=black},
ymin=-0.58, ymax=5.58
]
\draw[draw=none,fill=darkcyan1115178,fill opacity=0.75] (axis cs:-76.0900220327377,-0.3) rectangle (axis cs:-62.7650328969955,0.3);
\draw[draw=none,fill=darkorange2221435,fill opacity=0.75] (axis cs:-60.4493686628342,0.7) rectangle (axis cs:-52.6118664102554,1.3);
\draw[draw=none,fill=darkcyan2158115,fill opacity=0.75] (axis cs:-53.6115662593842,1.7) rectangle (axis cs:-50.8277456378937,2.3);
\draw[draw=none,fill=chocolate213940,fill opacity=0.75] (axis cs:-39.673004755497,2.7) rectangle (axis cs:-33.8933067359924,3.3);
\draw[draw=none,fill=orchid204120188,fill opacity=0.75] (axis cs:-33.159646894455,3.7) rectangle (axis cs:-29.9133306398392,4.3);
\draw[draw=none,fill=peru20214597,fill opacity=0.75] (axis cs:-52.5193189706802,4.7) rectangle (axis cs:-52.2459249725342,5.3);
\path [draw=black, draw opacity=0.5, semithick]
(axis cs:-68.9334267044067,-0.28125)
--(axis cs:-68.9334267044067,0.225);

\path [draw=black, draw opacity=0.5, semithick]
(axis cs:-56.1219903945923,0.71875)
--(axis cs:-56.1219903945923,1.225);

\path [draw=black, draw opacity=0.5, semithick]
(axis cs:-52.0833439254761,1.71875)
--(axis cs:-52.0833439254761,2.225);

\path [draw=black, draw opacity=0.5, semithick]
(axis cs:-36.7584313201904,2.71875)
--(axis cs:-36.7584313201904,3.225);

\path [draw=black, draw opacity=0.5, semithick]
(axis cs:-31.4792595672607,3.71875)
--(axis cs:-31.4792595672607,4.225);

\path [draw=black, draw opacity=0.5, semithick]
(axis cs:-52.3597022247314,4.71875)
--(axis cs:-52.3597022247314,5.225);

\end{groupplot}

\end{tikzpicture}

\input{pgf/Rliable/xyevolvev3perfprof}

\subsection{2D-Wind}
\begin{tikzpicture}

\definecolor{chocolate213940}{RGB}{213,94,0}
\definecolor{darkcyan1115178}{RGB}{1,115,178}
\definecolor{darkcyan2158115}{RGB}{2,158,115}
\definecolor{darkgray176}{RGB}{176,176,176}
\definecolor{darkorange2221435}{RGB}{222,143,5}
\definecolor{orchid204120188}{RGB}{204,120,188}
\definecolor{peru20214597}{RGB}{202,145,97}

\begin{groupplot}[
group style={
group size=3 by 1,
y descriptions at=edge left,
}, 
width=\linewidth/3.3, 
]\nextgroupplot[
tick align=outside,
tick pos=left,
title={Median},
x grid style={darkgray176},
xmajorgrids,
xmin=3.28999996930361, xmax=27.03506934506,
xtick style={color=black},
y grid style={darkgray176},
ymin=-0.58, ymax=5.58,
ytick style={color=black},
ytick={0,1,2,3,4,5},
yticklabels={VRNN,ContraBAR--,BOReL--,COSPA,Blind,Oracle}
]
\draw[draw=none,fill=darkcyan1115178,fill opacity=0.75] (axis cs:10.7216734665083,-0.3) rectangle (axis cs:16.2315849525301,0.3);
\addlegendimage{ybar,ybar legend,draw=none,fill=darkcyan1115178,fill opacity=0.75}

\draw[draw=none,fill=darkorange2221435,fill opacity=0.75] (axis cs:7.55999984741211,0.7) rectangle (axis cs:16.4699998378754,1.3);
\addlegendimage{ybar,ybar legend,draw=none,fill=darkorange2221435,fill opacity=0.75}

\draw[draw=none,fill=darkcyan2158115,fill opacity=0.75] (axis cs:8.25999977558851,1.7) rectangle (axis cs:12.059999537468,2.3);
\addlegendimage{ybar,ybar legend,draw=none,fill=darkcyan2158115,fill opacity=0.75}

\draw[draw=none,fill=chocolate213940,fill opacity=0.75] (axis cs:21.2813623049259,2.7) rectangle (axis cs:25.9043517557383,3.3);
\addlegendimage{ybar,ybar legend,draw=none,fill=chocolate213940,fill opacity=0.75}

\draw[draw=none,fill=orchid204120188,fill opacity=0.75] (axis cs:16.434999525547,3.7) rectangle (axis cs:21.2399993896484,4.3);
\addlegendimage{ybar,ybar legend,draw=none,fill=orchid204120188,fill opacity=0.75}

\draw[draw=none,fill=peru20214597,fill opacity=0.75] (axis cs:20.4299998521805,4.7) rectangle (axis cs:24.6349997520447,5.3);
\addlegendimage{ybar,ybar legend,draw=none,fill=peru20214597,fill opacity=0.75}

\path [draw=black, draw opacity=0.5, semithick]
(axis cs:13.5171808833024,-0.28125)
--(axis cs:13.5171808833024,0.225);

\path [draw=black, draw opacity=0.5, semithick]
(axis cs:11.9799998044968,0.71875)
--(axis cs:11.9799998044968,1.225);

\path [draw=black, draw opacity=0.5, semithick]
(axis cs:10.1399996638298,1.71875)
--(axis cs:10.1399996638298,2.225);

\path [draw=black, draw opacity=0.5, semithick]
(axis cs:23.7326650714874,2.71875)
--(axis cs:23.7326650714874,3.225);

\path [draw=black, draw opacity=0.5, semithick]
(axis cs:18.9449994087219,3.71875)
--(axis cs:18.9449994087219,4.225);

\path [draw=black, draw opacity=0.5, semithick]
(axis cs:22.7299997568131,4.71875)
--(axis cs:22.7299997568131,5.225);

\nextgroupplot[
tick align=outside,
tick pos=left,
title={IQM},
x grid style={darkgray176},
xmajorgrids,
xmin=0.169999992847443, xmax=27.541219910121,
xtick style={color=black},
ymin=-0.58, ymax=5.58
]
\draw[draw=none,fill=darkcyan1115178,fill opacity=0.75] (axis cs:10.0731452259254,-0.3) rectangle (axis cs:18.3432768671513,0.3);
\draw[draw=none,fill=darkorange2221435,fill opacity=0.75] (axis cs:4.55999994277954,0.7) rectangle (axis cs:18.4999998092651,1.3);
\draw[draw=none,fill=darkcyan2158115,fill opacity=0.75] (axis cs:7.85999984741211,1.7) rectangle (axis cs:12.2599994659424,2.3);
\draw[draw=none,fill=chocolate213940,fill opacity=0.75] (axis cs:24.2954735403061,2.7) rectangle (axis cs:26.2378284854889,3.3);
\draw[draw=none,fill=orchid204120188,fill opacity=0.75] (axis cs:17.8999992835522,3.7) rectangle (axis cs:23.5999994277954,4.3);
\draw[draw=none,fill=peru20214597,fill opacity=0.75] (axis cs:24.3199995040894,4.7) rectangle (axis cs:25.9399997711182,5.3);
\path [draw=black, draw opacity=0.5, semithick]
(axis cs:14.5233791685104,-0.28125)
--(axis cs:14.5233791685104,0.225);

\path [draw=black, draw opacity=0.5, semithick]
(axis cs:11.1199998378754,0.71875)
--(axis cs:11.1199998378754,1.225);

\path [draw=black, draw opacity=0.5, semithick]
(axis cs:10.2199995994568,1.71875)
--(axis cs:10.2199995994568,2.225);

\path [draw=black, draw opacity=0.5, semithick]
(axis cs:25.8257579994202,2.71875)
--(axis cs:25.8257579994202,3.225);

\path [draw=black, draw opacity=0.5, semithick]
(axis cs:21.2199994087219,3.71875)
--(axis cs:21.2199994087219,4.225);

\path [draw=black, draw opacity=0.5, semithick]
(axis cs:25.7799995422363,4.71875)
--(axis cs:25.7799995422363,5.225);

\nextgroupplot[
tick align=outside,
tick pos=left,
title={Mean},
x grid style={darkgray176},
xmajorgrids,
xmin=3.28999996930361, xmax=27.03506934506,
xtick style={color=black},
ymin=-0.58, ymax=5.58
]
\draw[draw=none,fill=darkcyan1115178,fill opacity=0.75] (axis cs:10.7216734665083,-0.3) rectangle (axis cs:16.2315849525301,0.3);
\draw[draw=none,fill=darkorange2221435,fill opacity=0.75] (axis cs:7.55999984741211,0.7) rectangle (axis cs:16.4699998378754,1.3);
\draw[draw=none,fill=darkcyan2158115,fill opacity=0.75] (axis cs:8.25999977558851,1.7) rectangle (axis cs:12.059999537468,2.3);
\draw[draw=none,fill=chocolate213940,fill opacity=0.75] (axis cs:21.2813623049259,2.7) rectangle (axis cs:25.9043517557383,3.3);
\draw[draw=none,fill=orchid204120188,fill opacity=0.75] (axis cs:16.434999525547,3.7) rectangle (axis cs:21.2399993896484,4.3);
\draw[draw=none,fill=peru20214597,fill opacity=0.75] (axis cs:20.4299998521805,4.7) rectangle (axis cs:24.6349997520447,5.3);
\path [draw=black, draw opacity=0.5, semithick]
(axis cs:13.5171808833024,-0.28125)
--(axis cs:13.5171808833024,0.225);

\path [draw=black, draw opacity=0.5, semithick]
(axis cs:11.9799998044968,0.71875)
--(axis cs:11.9799998044968,1.225);

\path [draw=black, draw opacity=0.5, semithick]
(axis cs:10.1399996638298,1.71875)
--(axis cs:10.1399996638298,2.225);

\path [draw=black, draw opacity=0.5, semithick]
(axis cs:23.7326650714874,2.71875)
--(axis cs:23.7326650714874,3.225);

\path [draw=black, draw opacity=0.5, semithick]
(axis cs:18.9449994087219,3.71875)
--(axis cs:18.9449994087219,4.225);

\path [draw=black, draw opacity=0.5, semithick]
(axis cs:22.7299997568131,4.71875)
--(axis cs:22.7299997568131,5.225);

\end{groupplot}

\end{tikzpicture}

\input{pgf/Rliable/xywindperfprof}

\subsection{Ant-Weight}
\begin{tikzpicture}

\definecolor{chocolate213940}{RGB}{213,94,0}
\definecolor{darkcyan1115178}{RGB}{1,115,178}
\definecolor{darkcyan2158115}{RGB}{2,158,115}
\definecolor{darkgray176}{RGB}{176,176,176}
\definecolor{darkorange2221435}{RGB}{222,143,5}
\definecolor{orchid204120188}{RGB}{204,120,188}
\definecolor{peru20214597}{RGB}{202,145,97}

\begin{groupplot}[
group style={
group size=3 by 1,
y descriptions at=edge left,
}, 
width=\linewidth/3.3, 
scaled x ticks = true,
scaled x ticks=base 10:-3,
]
\nextgroupplot[
tick align=outside,
tick pos=left,
title={Median},
x grid style={darkgray176},
xmajorgrids,
xmin=2395.61007064819, xmax=3289.77834783783,
xtick style={color=black},
y grid style={darkgray176},
ymin=-0.58, ymax=5.58,
ytick style={color=black},
ytick={0,1,2,3,4,5},
yticklabels={VRNN,ContraBAR,BOREL,COSPA,Blind,Oracle}
]
\draw[draw=none,fill=darkcyan1115178,fill opacity=0.75] (axis cs:2756.9314289093,-0.3) rectangle (axis cs:3247.19890606689,0.3);
\addlegendimage{ybar,ybar legend,draw=none,fill=darkcyan1115178,fill opacity=0.75}

\draw[draw=none,fill=darkorange2221435,fill opacity=0.75] (axis cs:2635.06589974976,0.7) rectangle (axis cs:2790.86609790039,1.3);
\addlegendimage{ybar,ybar legend,draw=none,fill=darkorange2221435,fill opacity=0.75}

\draw[draw=none,fill=darkcyan2158115,fill opacity=0.75] (axis cs:2395.61007064819,1.7) rectangle (axis cs:2824.84560963949,2.3);
\addlegendimage{ybar,ybar legend,draw=none,fill=darkcyan2158115,fill opacity=0.75}

\draw[draw=none,fill=chocolate213940,fill opacity=0.75] (axis cs:3003.97011755371,2.7) rectangle (axis cs:3203.0384833374,3.3);
\addlegendimage{ybar,ybar legend,draw=none,fill=chocolate213940,fill opacity=0.75}

\draw[draw=none,fill=orchid204120188,fill opacity=0.75] (axis cs:2693.91690785375,3.7) rectangle (axis cs:2888.16762813174,4.3);
\addlegendimage{ybar,ybar legend,draw=none,fill=orchid204120188,fill opacity=0.75}

\draw[draw=none,fill=peru20214597,fill opacity=0.75] (axis cs:2652.11220157021,4.7) rectangle (axis cs:2841.13853971783,5.3);
\addlegendimage{ybar,ybar legend,draw=none,fill=peru20214597,fill opacity=0.75}

\path [draw=black, draw opacity=0.5, semithick]
(axis cs:3035.36538024902,-0.28125)
--(axis cs:3035.36538024902,0.225);

\path [draw=black, draw opacity=0.5, semithick]
(axis cs:2715.13345031738,0.71875)
--(axis cs:2715.13345031738,1.225);

\path [draw=black, draw opacity=0.5, semithick]
(axis cs:2637.45844014486,1.71875)
--(axis cs:2637.45844014486,2.225);

\path [draw=black, draw opacity=0.5, semithick]
(axis cs:3104.39741943359,2.71875)
--(axis cs:3104.39741943359,3.225);

\path [draw=black, draw opacity=0.5, semithick]
(axis cs:2797.27484762258,3.71875)
--(axis cs:2797.27484762258,4.225);

\path [draw=black, draw opacity=0.5, semithick]
(axis cs:2750.1855417352,4.71875)
--(axis cs:2750.1855417352,5.225);

\nextgroupplot[
tick align=outside,
tick pos=left,
title={IQM},
x grid style={darkgray176},
xmajorgrids,
xmin=2645.06195170085, xmax=3362.56599840495,
xtick style={color=black},
ymin=-0.58, ymax=5.58
]
\draw[draw=none,fill=darkcyan1115178,fill opacity=0.75] (axis cs:2961.77802648926,-0.3) rectangle (axis cs:3328.39913903809,0.3);
\draw[draw=none,fill=darkorange2221435,fill opacity=0.75] (axis cs:2657.35957397461,0.7) rectangle (axis cs:2824.84599273682,1.3);
\draw[draw=none,fill=darkcyan2158115,fill opacity=0.75] (axis cs:2645.06195170085,1.7) rectangle (axis cs:2891.92045013428,2.3);
\draw[draw=none,fill=chocolate213940,fill opacity=0.75] (axis cs:3017.32826928711,2.7) rectangle (axis cs:3213.33557128906,3.3);
\draw[draw=none,fill=orchid204120188,fill opacity=0.75] (axis cs:2749.78640226237,3.7) rectangle (axis cs:2921.20044238281,4.3);
\draw[draw=none,fill=peru20214597,fill opacity=0.75] (axis cs:2682.61092318448,4.7) rectangle (axis cs:2843.32758023349,5.3);
\path [draw=black, draw opacity=0.5, semithick]
(axis cs:3207.72295898437,-0.28125)
--(axis cs:3207.72295898437,0.225);

\path [draw=black, draw opacity=0.5, semithick]
(axis cs:2746.62879882813,0.71875)
--(axis cs:2746.62879882813,1.225);

\path [draw=black, draw opacity=0.5, semithick]
(axis cs:2802.22171630859,1.71875)
--(axis cs:2802.22171630859,2.225);

\path [draw=black, draw opacity=0.5, semithick]
(axis cs:3090.40220703125,2.71875)
--(axis cs:3090.40220703125,3.225);

\path [draw=black, draw opacity=0.5, semithick]
(axis cs:2845.27840820313,3.71875)
--(axis cs:2845.27840820313,4.225);

\path [draw=black, draw opacity=0.5, semithick]
(axis cs:2772.22691761364,4.71875)
--(axis cs:2772.22691761364,5.225);

\nextgroupplot[
tick align=outside,
tick pos=left,
title={Mean},
x grid style={darkgray176},
xmajorgrids,
xmin=2395.61007064819, xmax=3289.77834783783,
xtick style={color=black},
ymin=-0.58, ymax=5.58
]
\draw[draw=none,fill=darkcyan1115178,fill opacity=0.75] (axis cs:2756.9314289093,-0.3) rectangle (axis cs:3247.19890606689,0.3);
\draw[draw=none,fill=darkorange2221435,fill opacity=0.75] (axis cs:2635.06589974976,0.7) rectangle (axis cs:2790.86609790039,1.3);
\draw[draw=none,fill=darkcyan2158115,fill opacity=0.75] (axis cs:2395.61007064819,1.7) rectangle (axis cs:2824.84560963949,2.3);
\draw[draw=none,fill=chocolate213940,fill opacity=0.75] (axis cs:3003.97011755371,2.7) rectangle (axis cs:3203.0384833374,3.3);
\draw[draw=none,fill=orchid204120188,fill opacity=0.75] (axis cs:2693.91690785375,3.7) rectangle (axis cs:2888.16762813174,4.3);
\draw[draw=none,fill=peru20214597,fill opacity=0.75] (axis cs:2652.11220157021,4.7) rectangle (axis cs:2841.13853971783,5.3);
\path [draw=black, draw opacity=0.5, semithick]
(axis cs:3035.36538024902,-0.28125)
--(axis cs:3035.36538024902,0.225);

\path [draw=black, draw opacity=0.5, semithick]
(axis cs:2715.13345031738,0.71875)
--(axis cs:2715.13345031738,1.225);

\path [draw=black, draw opacity=0.5, semithick]
(axis cs:2637.45844014486,1.71875)
--(axis cs:2637.45844014486,2.225);

\path [draw=black, draw opacity=0.5, semithick]
(axis cs:3104.39741943359,2.71875)
--(axis cs:3104.39741943359,3.225);

\path [draw=black, draw opacity=0.5, semithick]
(axis cs:2797.27484762258,3.71875)
--(axis cs:2797.27484762258,4.225);

\path [draw=black, draw opacity=0.5, semithick]
(axis cs:2750.1855417352,4.71875)
--(axis cs:2750.1855417352,5.225);

\end{groupplot}

\end{tikzpicture}

\input{pgf/Rliable/antweightperfprof}

\subsection{Ant-Leg}
\begin{tikzpicture}

\definecolor{chocolate213940}{RGB}{213,94,0}
\definecolor{darkcyan1115178}{RGB}{1,115,178}
\definecolor{darkcyan2158115}{RGB}{2,158,115}
\definecolor{darkgray176}{RGB}{176,176,176}
\definecolor{darkorange2221435}{RGB}{222,143,5}
\definecolor{orchid204120188}{RGB}{204,120,188}
\definecolor{peru20214597}{RGB}{202,145,97}

\begin{groupplot}[
group style={
group size=3 by 1,
y descriptions at=edge left,
}, 
width=\linewidth/3.3, 
scaled x ticks = true,
scaled x ticks=base 10:-3,
]
\nextgroupplot[
tick align=outside,
tick pos=left,
title={Median},
x grid style={darkgray176},
xmajorgrids,
xmin=645.040318804654, xmax=1600.81945446687,
xtick style={color=black},
y grid style={darkgray176},
ymin=-0.58, ymax=5.58,
ytick style={color=black},
ytick={0,1,2,3,4,5},
yticklabels={VRNN,ContraBAR,COSPA,BOReL,Oracle,Blind}
]
\draw[draw=none,fill=darkcyan1115178,fill opacity=0.75] (axis cs:1133.05789570236,-0.3) rectangle (axis cs:1356.34016300964,0.3);
\addlegendimage{ybar,ybar legend,draw=none,fill=darkcyan1115178,fill opacity=0.75}

\draw[draw=none,fill=darkorange2221435,fill opacity=0.75] (axis cs:1023.11875804901,0.7) rectangle (axis cs:1342.78380813599,1.3);
\addlegendimage{ybar,ybar legend,draw=none,fill=darkorange2221435,fill opacity=0.75}

\draw[draw=none,fill=darkcyan2158115,fill opacity=0.75] (axis cs:1303.89426641846,1.7) rectangle (axis cs:1555.30616229248,2.3);
\addlegendimage{ybar,ybar legend,draw=none,fill=darkcyan2158115,fill opacity=0.75}

\draw[draw=none,fill=chocolate213940,fill opacity=0.75] (axis cs:645.040318804654,2.7) rectangle (axis cs:943.367540820728,3.3);
\addlegendimage{ybar,ybar legend,draw=none,fill=chocolate213940,fill opacity=0.75}

\draw[draw=none,fill=orchid204120188,fill opacity=0.75] (axis cs:1233.5470909236,3.7) rectangle (axis cs:1515.47916391226,4.3);
\addlegendimage{ybar,ybar legend,draw=none,fill=orchid204120188,fill opacity=0.75}

\draw[draw=none,fill=peru20214597,fill opacity=0.75] (axis cs:1105.08106204693,4.7) rectangle (axis cs:1425.07226928711,5.3);
\addlegendimage{ybar,ybar legend,draw=none,fill=peru20214597,fill opacity=0.75}

\path [draw=black, draw opacity=0.5, semithick]
(axis cs:1260.62371170044,-0.28125)
--(axis cs:1260.62371170044,0.225);

\path [draw=black, draw opacity=0.5, semithick]
(axis cs:1194.75359588623,0.71875)
--(axis cs:1194.75359588623,1.225);

\path [draw=black, draw opacity=0.5, semithick]
(axis cs:1441.50832275391,1.71875)
--(axis cs:1441.50832275391,2.225);

\path [draw=black, draw opacity=0.5, semithick]
(axis cs:795.448024194891,2.71875)
--(axis cs:795.448024194891,3.225);

\path [draw=black, draw opacity=0.5, semithick]
(axis cs:1380.29137901893,3.71875)
--(axis cs:1380.29137901893,4.225);

\path [draw=black, draw opacity=0.5, semithick]
(axis cs:1287.36309767503,4.71875)
--(axis cs:1287.36309767503,5.225);

\nextgroupplot[
tick align=outside,
tick pos=left,
title={IQM},
x grid style={darkgray176},
xmajorgrids,
xmin=640.743837604523, xmax=1652.79619126797,
xtick style={color=black},
ymin=-0.58, ymax=5.58,
]
\draw[draw=none,fill=darkcyan1115178,fill opacity=0.75] (axis cs:1232.28293307495,-0.3) rectangle (axis cs:1388.12775964355,0.3);
\draw[draw=none,fill=darkorange2221435,fill opacity=0.75] (axis cs:1100.64058242798,0.7) rectangle (axis cs:1414.33363262939,1.3);
\draw[draw=none,fill=darkcyan2158115,fill opacity=0.75] (axis cs:1399.04730322266,1.7) rectangle (axis cs:1604.6032220459,2.3);
\draw[draw=none,fill=chocolate213940,fill opacity=0.75] (axis cs:640.743837604523,2.7) rectangle (axis cs:975.476611735026,3.3);
\draw[draw=none,fill=orchid204120188,fill opacity=0.75] (axis cs:1207.58415178571,3.7) rectangle (axis cs:1584.50393589565,4.3);
\draw[draw=none,fill=peru20214597,fill opacity=0.75] (axis cs:1210.8943737139,4.7) rectangle (axis cs:1448.04568777902,5.3);
\path [draw=black, draw opacity=0.5, semithick]
(axis cs:1325.12026367188,-0.28125)
--(axis cs:1325.12026367188,0.225);

\path [draw=black, draw opacity=0.5, semithick]
(axis cs:1324.56230957031,0.71875)
--(axis cs:1324.56230957031,1.225);

\path [draw=black, draw opacity=0.5, semithick]
(axis cs:1528.17936035156,1.71875)
--(axis cs:1528.17936035156,2.225);

\path [draw=black, draw opacity=0.5, semithick]
(axis cs:811.487956746419,2.71875)
--(axis cs:811.487956746419,3.225);

\path [draw=black, draw opacity=0.5, semithick]
(axis cs:1428.96432407924,3.71875)
--(axis cs:1428.96432407924,4.225);

\path [draw=black, draw opacity=0.5, semithick]
(axis cs:1396.75091727121,4.71875)
--(axis cs:1396.75091727121,5.225);

\nextgroupplot[
tick align=outside,
tick pos=left,
title={Mean},
x grid style={darkgray176},
xmajorgrids,
xmin=645.040318804654, xmax=1600.81945446687,
xtick style={color=black},
ymin=-0.58, ymax=5.58
]
\draw[draw=none,fill=darkcyan1115178,fill opacity=0.75] (axis cs:1133.05789570236,-0.3) rectangle (axis cs:1356.34016300964,0.3);
\draw[draw=none,fill=darkorange2221435,fill opacity=0.75] (axis cs:1023.11875804901,0.7) rectangle (axis cs:1342.78380813599,1.3);
\draw[draw=none,fill=darkcyan2158115,fill opacity=0.75] (axis cs:1303.89426641846,1.7) rectangle (axis cs:1555.30616229248,2.3);
\draw[draw=none,fill=chocolate213940,fill opacity=0.75] (axis cs:645.040318804654,2.7) rectangle (axis cs:943.367540820728,3.3);
\draw[draw=none,fill=orchid204120188,fill opacity=0.75] (axis cs:1233.5470909236,3.7) rectangle (axis cs:1515.47916391226,4.3);
\draw[draw=none,fill=peru20214597,fill opacity=0.75] (axis cs:1105.08106204693,4.7) rectangle (axis cs:1425.07226928711,5.3);
\path [draw=black, draw opacity=0.5, semithick]
(axis cs:1260.62371170044,-0.28125)
--(axis cs:1260.62371170044,0.225);

\path [draw=black, draw opacity=0.5, semithick]
(axis cs:1194.75359588623,0.71875)
--(axis cs:1194.75359588623,1.225);

\path [draw=black, draw opacity=0.5, semithick]
(axis cs:1441.50832275391,1.71875)
--(axis cs:1441.50832275391,2.225);

\path [draw=black, draw opacity=0.5, semithick]
(axis cs:795.448024194891,2.71875)
--(axis cs:795.448024194891,3.225);

\path [draw=black, draw opacity=0.5, semithick]
(axis cs:1380.29137901893,3.71875)
--(axis cs:1380.29137901893,4.225);

\path [draw=black, draw opacity=0.5, semithick]
(axis cs:1287.36309767503,4.71875)
--(axis cs:1287.36309767503,5.225);

\end{groupplot}

\end{tikzpicture}

\input{pgf/Rliable/antleglenperfprof}

\subsection{Barkour-Weight}
\begin{tikzpicture}

\definecolor{chocolate213940}{RGB}{213,94,0}
\definecolor{darkcyan1115178}{RGB}{1,115,178}
\definecolor{darkcyan2158115}{RGB}{2,158,115}
\definecolor{darkgray176}{RGB}{176,176,176}
\definecolor{darkorange2221435}{RGB}{222,143,5}
\definecolor{orchid204120188}{RGB}{204,120,188}
\definecolor{peru20214597}{RGB}{202,145,97}

\begin{groupplot}[
group style={
group size=3 by 1,
y descriptions at=edge left,
}, 
width=\linewidth/3.3, 
]
\nextgroupplot[
tick align=outside,
tick pos=left,
title={Median},
x grid style={darkgray176},
xmajorgrids,
xmin=13.3772967243195, xmax=19.0238837795258,
xtick style={color=black},
y grid style={darkgray176},
ymin=-0.58, ymax=5.58,
ytick style={color=black},
ytick={0,1,2,3,4,5},
yticklabels={ContraBAR,VRNN,Blind,COSPA,Oracle,BOReL}
]
\draw[draw=none,fill=darkcyan1115178,fill opacity=0.75] (axis cs:13.3772967243195,-0.3) rectangle (axis cs:16.874438624382,0.3);
\addlegendimage{ybar,ybar legend,draw=none,fill=darkcyan1115178,fill opacity=0.75}

\draw[draw=none,fill=darkorange2221435,fill opacity=0.75] (axis cs:15.5590076225996,0.7) rectangle (axis cs:16.5111319690943,1.3);
\addlegendimage{ybar,ybar legend,draw=none,fill=darkorange2221435,fill opacity=0.75}

\draw[draw=none,fill=darkcyan2158115,fill opacity=0.75] (axis cs:16.958495990634,1.7) rectangle (axis cs:17.8276377302408,2.3);
\addlegendimage{ybar,ybar legend,draw=none,fill=darkcyan2158115,fill opacity=0.75}

\draw[draw=none,fill=chocolate213940,fill opacity=0.75] (axis cs:17.7425426971912,2.7) rectangle (axis cs:18.2822159290314,3.3);
\addlegendimage{ybar,ybar legend,draw=none,fill=chocolate213940,fill opacity=0.75}

\draw[draw=none,fill=orchid204120188,fill opacity=0.75] (axis cs:18.5300460270473,3.7) rectangle (axis cs:18.7549986816588,4.3);
\addlegendimage{ybar,ybar legend,draw=none,fill=orchid204120188,fill opacity=0.75}

\draw[draw=none,fill=peru20214597,fill opacity=0.75] (axis cs:16.9943767094612,4.7) rectangle (axis cs:18.0241304302216,5.3);
\addlegendimage{ybar,ybar legend,draw=none,fill=peru20214597,fill opacity=0.75}

\path [draw=black, draw opacity=0.5, semithick]
(axis cs:15.2541897296906,-0.28125)
--(axis cs:15.2541897296906,0.225);

\path [draw=black, draw opacity=0.5, semithick]
(axis cs:16.0484705448151,0.71875)
--(axis cs:16.0484705448151,1.225);

\path [draw=black, draw opacity=0.5, semithick]
(axis cs:17.410667681694,1.71875)
--(axis cs:17.410667681694,2.225);

\path [draw=black, draw opacity=0.5, semithick]
(axis cs:18.0515351772308,2.71875)
--(axis cs:18.0515351772308,3.225);

\path [draw=black, draw opacity=0.5, semithick]
(axis cs:18.6542049589611,3.71875)
--(axis cs:18.6542049589611,4.225);

\path [draw=black, draw opacity=0.5, semithick]
(axis cs:17.5266649723053,4.71875)
--(axis cs:17.5266649723053,5.225);

\nextgroupplot[
tick align=outside,
tick pos=left,
title={IQM},
x grid style={darkgray176},
xmajorgrids,
xmin=14.4577572131157, xmax=19.0078404196718,
xtick style={color=black},
ymin=-0.58, ymax=5.58
]
\draw[draw=none,fill=darkcyan1115178,fill opacity=0.75] (axis cs:14.4577572131157,-0.3) rectangle (axis cs:17.7871683979034,0.3);
\draw[draw=none,fill=darkorange2221435,fill opacity=0.75] (axis cs:15.6909257411957,0.7) rectangle (axis cs:16.7147241222858,1.3);
\draw[draw=none,fill=darkcyan2158115,fill opacity=0.75] (axis cs:17.2012563061714,1.7) rectangle (axis cs:18.1521712827682,2.3);
\draw[draw=none,fill=chocolate213940,fill opacity=0.75] (axis cs:18.016139087677,2.7) rectangle (axis cs:18.3222236633301,3.3);
\draw[draw=none,fill=orchid204120188,fill opacity=0.75] (axis cs:18.6264737779444,3.7) rectangle (axis cs:18.7911697907881,4.3);
\draw[draw=none,fill=peru20214597,fill opacity=0.75] (axis cs:16.9230749893188,4.7) rectangle (axis cs:18.3458431243896,5.3);
\path [draw=black, draw opacity=0.5, semithick]
(axis cs:16.5321494102478,-0.28125)
--(axis cs:16.5321494102478,0.225);

\path [draw=black, draw opacity=0.5, semithick]
(axis cs:16.1720916748047,0.71875)
--(axis cs:16.1720916748047,1.225);

\path [draw=black, draw opacity=0.5, semithick]
(axis cs:17.7323544502258,1.71875)
--(axis cs:17.7323544502258,2.225);

\path [draw=black, draw opacity=0.5, semithick]
(axis cs:18.1723262786865,2.71875)
--(axis cs:18.1723262786865,3.225);

\path [draw=black, draw opacity=0.5, semithick]
(axis cs:18.7100696563721,3.71875)
--(axis cs:18.7100696563721,4.225);

\path [draw=black, draw opacity=0.5, semithick]
(axis cs:17.6803813934326,4.71875)
--(axis cs:17.6803813934326,5.225);

\nextgroupplot[
tick align=outside,
tick pos=left,
title={Mean},
x grid style={darkgray176},
xmajorgrids,
xmin=13.3772967243195, xmax=19.0238837795258,
xtick style={color=black},
ymin=-0.58, ymax=5.58
]
\draw[draw=none,fill=darkcyan1115178,fill opacity=0.75] (axis cs:13.3772967243195,-0.3) rectangle (axis cs:16.874438624382,0.3);
\draw[draw=none,fill=darkorange2221435,fill opacity=0.75] (axis cs:15.5590076225996,0.7) rectangle (axis cs:16.5111319690943,1.3);
\draw[draw=none,fill=darkcyan2158115,fill opacity=0.75] (axis cs:16.958495990634,1.7) rectangle (axis cs:17.8276377302408,2.3);
\draw[draw=none,fill=chocolate213940,fill opacity=0.75] (axis cs:17.7425426971912,2.7) rectangle (axis cs:18.2822159290314,3.3);
\draw[draw=none,fill=orchid204120188,fill opacity=0.75] (axis cs:18.5300460270473,3.7) rectangle (axis cs:18.7549986816588,4.3);
\draw[draw=none,fill=peru20214597,fill opacity=0.75] (axis cs:16.9943767094612,4.7) rectangle (axis cs:18.0241304302216,5.3);
\path [draw=black, draw opacity=0.5, semithick]
(axis cs:15.2541897296906,-0.28125)
--(axis cs:15.2541897296906,0.225);

\path [draw=black, draw opacity=0.5, semithick]
(axis cs:16.0484705448151,0.71875)
--(axis cs:16.0484705448151,1.225);

\path [draw=black, draw opacity=0.5, semithick]
(axis cs:17.410667681694,1.71875)
--(axis cs:17.410667681694,2.225);

\path [draw=black, draw opacity=0.5, semithick]
(axis cs:18.0515351772308,2.71875)
--(axis cs:18.0515351772308,3.225);

\path [draw=black, draw opacity=0.5, semithick]
(axis cs:18.6542049589611,3.71875)
--(axis cs:18.6542049589611,4.225);

\path [draw=black, draw opacity=0.5, semithick]
(axis cs:17.5266649723053,4.71875)
--(axis cs:17.5266649723053,5.225);

\end{groupplot}

\end{tikzpicture}

\input{pgf/Rliable/bark-weight-noangrew-perfprof}

\end{document}